\documentclass[preprint,12pt,numbers,sort&compress]{elsarticle}
\usepackage{amssymb}
\usepackage{amsmath}
\usepackage{verbatim}
\usepackage{graphics}
\usepackage{algorithm}
\usepackage{algorithmic}
\usepackage{graphicx}
\usepackage{epstopdf}
\usepackage{subfigure}
\usepackage{setspace}
\usepackage{float}
\usepackage{multirow}
\usepackage{stfloats}
\usepackage{array}
\usepackage{multirow}
\usepackage{footmisc}
\usepackage{threeparttable}
\usepackage{caption}
\journal{Neurocomputing}

\begin{document}
\captionsetup[figure]{labelfont={bf},labelformat={default},labelsep=period,name={Fig.}}
\captionsetup[table]{labelfont={bf},labelformat={default},labelsep=none,name={Table}}
\begin{frontmatter}

\title{CISRDCNN: Super-resolution of compressed images using deep convolutional neural networks}

\author{Honggang Chen}
\ead{honggang.chen@stu.scu.edu.cn}
\author{Xiaohai He\corref{cor}}
\ead{hxh@scu.edu.cn}
\author{Chao Ren\corref{}}
\ead{chaoren@scu.edu.cn}
\author{Linbo Qing\corref{}}
\ead{qing\_lb@scu.edu.cn}
\author{Qizhi Teng\corref{}}
\ead{qzteng@scu.edu.cn}

\cortext[cor]{Corresponding author.}
\address{College of Electronics and Information Engineering, Sichuan University, Chengdu, China}

\begin{abstract}

In recent years, much research has been conducted on image super-resolution (SR). To the best of our knowledge, however, few SR methods were concerned with compressed images. The SR of compressed images is a challenging task due to the complicated compression artifacts, while many images suffer from them in practice. The intuitive solution for this difficult task is to decouple it into two sequential but independent subproblems, i.e., compression artifacts reduction (CAR) and SR. Nevertheless, some useful details may be removed in CAR stage, which is contrary to the goal of SR and makes the SR stage more challenging. In this paper, an end-to-end trainable deep convolutional neural network is designed to perform SR on compressed images (CISRDCNN), which reduces compression artifacts and improves image resolution jointly. Experiments on compressed images produced by JPEG (we take the JPEG as an example in this paper) demonstrate that the proposed CISRDCNN yields state-of-the-art SR performance on commonly used test images and imagesets. The results of CISRDCNN on real low quality web images are also very impressive, with obvious quality enhancement. Further, we explore the application of the proposed SR method in low bit-rate image coding, leading to better rate-distortion performance than JPEG.
\end{abstract}


\end{frontmatter}

\section{Introduction}
 Single image super-resolution (SISR) refers to estimate a high-resolution (HR) image from a single low-resolution (LR) observation, which is of great significance to many image processing and analysis systems. However, the SISR problem is very challenging duo to the ill-posed condition. In other words, a LR image corresponds to a set of HR images, while most of them are not expected. In general, the reconstructed HR image should be visually pleasant and close to the real one as much as possible.

The SISR problem has been widely researched over the past 20 years and plenty of algorithms have been proposed. Roughly speaking, interpolation-based \cite{1,2,3,4,5,6,7,8,9}, reconstruction-based \cite{10,11,12,13,14,15,16,17,18,19,20}, and learning-based \cite{21,22,23,24,25,26,27,28,29,30,31,32,33,34,35,36,37,38,39,40,41,42,43} algorithms are the three main classes of SISR methods. Generally, the interpolation-based super-resolution (SR) approaches estimate unknown HR pixels using their neighborhoods (the known LR pixels) according to local structure properties. For the reconstruction-based methods, the observation model of LR image and prior knowledge of HR image are integrated to formulate an energy function, and thus the SR task can be converted to an optimization problem. The prior knowledge, which greatly affects SR performance, is the research focus for this kind of methods. The commonly used priors include gradient \cite{10,11,12}, sparsity \cite{13,14,15}, nonlocal self-similarity \cite{14,15,16,17,18,19,20}, etc. Many reconstruction-based SR methods use two or more priors to combine their complementary properties. The pre-trained mapping between LR images and HR images is usually adopted to guide the SR process in learning-based methods. According to the core of learning-based methods, it can be roughly divided into the following five subclasses further, i.e., neighbor embedding-based \cite{21,22,23}, example-based \cite{24,25,26,27}, sparse coding-based \cite{28,29,30,31,32}, regression-based \cite{33,34,35,36}, and deep learning-based \cite{37,38,39,40,41,42,43}. With fast execution speed and outstanding restoration quality, deep learning-based methods show great potential for SR problem. Meanwhile, some researchers attempted to combine different kinds of SR methods, thus integrating their merits \cite{44,45}.

In some practical applications, such as mobile communication and internet, limited by storage capacity and transmission bandwidth, images and videos are generally downsampled and compressed to reduce data volume. In these cases, the observations usually suffer from both of the downsampling and compression degradations, which makes the SR problem more difficult. Although much research has been done on SISR problem and plenty of effective SISR methods have been proposed over the past few decades, few methods were concerned with compressed images \cite{46,47,48,49,50,51}. Roughly, there are two kinds of frameworks for compressed images SR. Some researchers converted this task to an optimization problem via compression process modeling and prior knowledge regularization. In SRCDFOE \cite{46}, the compression distortion is seen as the spatially correlated Gaussian noise, and the Markov random field and total variation are used to regularize the estimated HR images. To realize decompression and SR simultaneously, the DCSRMOTV \cite{47} incorporates multi-order total variation model into JPEG image acquisition model. For this type of methods, the main pain point is how to realize the accurate modeling of compression process. In addition, it is difficult to balance compression artifacts reduction (CAR) and details preservation. Another commonly used strategy is to decompose this task into two subproblems (i.e., CAR and SR) and use a cascading framework to address them. For example, Xiong et al. \cite{48} combined adaptive regularization and learning-based SR to reduce compression noise and compensate details, respectively. Kang et al. \cite{49} proposed a sparse coding-based SR method for compressed images, in which the patches with/without compression artifacts are processed differently. Using a denoised training dataset, Lee et al. \cite{50} presented a dual-learning-based algorithm for compression noise reduction and SR. More recently, Zhao et al. \cite{51} constructed a three-steps-process framework for compressed images SR, which is composed of BM3D filtering-based compression noise reduction, local encoding-based patch classification, and mapping-based reconstruction. However, for most of this kind of algorithms, the compression noise reduction and upsampling are treated as two independent stages. Consequently, the resultant images of existing methods are apt to still contain compression noise or be over smoothed. On the whole, the research on SR of compressed images is lacking and there is still much room for performance improvement.

\begin{figure}[!tb] 
\centering
\subfigure[]{
\includegraphics[width=6cm]{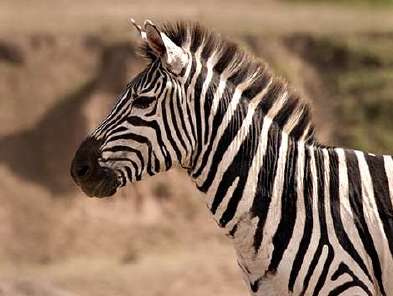}}
\subfigure[]{
\includegraphics[width=6cm]{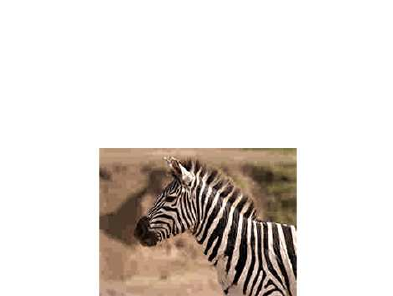}}\\
\subfigure[]{
\includegraphics[width=6cm]{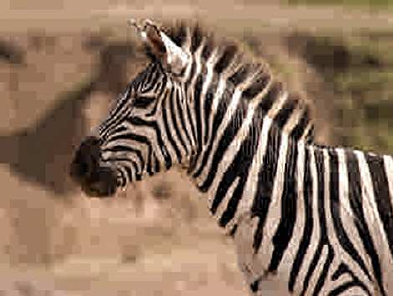}}
\subfigure[]{
\includegraphics[width=6cm]{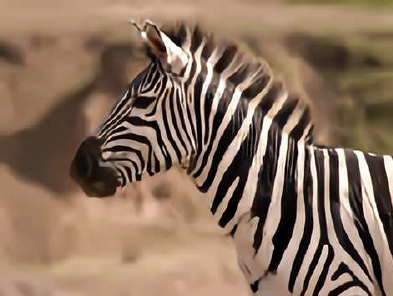}}
\caption{Illustration for JPEG compressed image SR on test image \emph{Zebra} (SR factor: 2, QF: 10). (a) Original image. (b) JPEG compressed LR image. (c) Result of Bicubic on (b). (d) Result of CISRDCNN on (b). Obviously, our result (d) is more visually pleasant than (b) and (c). Please zoom in to view details and make comparisons.}
\label{figure.1}
\end{figure}

\begin{center}
\begin{figure*}[!tb]
\centering
\includegraphics{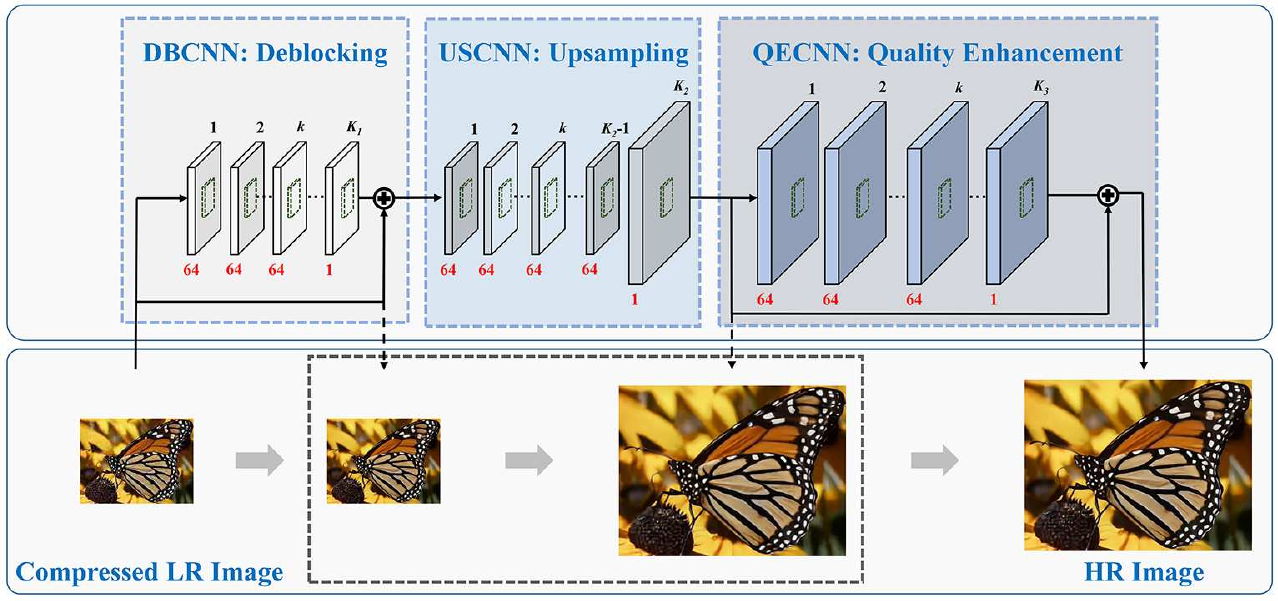}
\caption{The flowchart of CISRDCNN. Top: the architecture of CISRDCNN. Bottom: the illustration of reconstruction process.}
\label{figure.2}
\end{figure*}
\end{center}

The core issue of compressed images SR is how to reduce compression noise and preserve details as much as possible when enhancing image resolution. On the one hand, it is hard to remove compression artifacts in super-resolved images without a CAR or denoising stage. What is worse, the compression noise in LR may be significantly magnified in HR. On the other hand, the CAR and SR operations should not be separated as part of the details removed in CAR stage are useful for SR. On the basis of the above insights, an end-to-end trainable deep convolutional neural network (CNN) is designed to perform SR on compressed images, and we name it CISRDCNN. The CISRDCNN takes the compressed LR image as input and outputs the resultant HR image directly, without any preprocessing or postprocessing stage. Fig. \ref{figure.1} gives an example of the result of CISRDCNN, and we can see that our result is much more visually pleasant than the LR input and the resultant image of Bicubic interpolation. The framework of the proposed CISRDCNN is illustrated in Fig. \ref{figure.2}, and our contributions in this work are mainly in the following aspects:
\begin{itemize}

\item We propose a deep CNN-based SR framework for compressed images, which reduces compression artifacts and enhances image resolution simultaneously.

\item To preserve the functions of different modules in CISRDCNN and achieve joint optimization of CAR and SR, a special strategy is used to train the proposed network, i.e., individual training and joint optimization.

\item Extensive experiments show that the proposed CISRDCNN achieves outstanding SR performance on simulation experiments as well as the test on real low quality web images.

\item We explore the application of the proposed CISRDCNN in low bit-rate image coding, and the experimental results demonstrate that it can improve the rate-distortion performance of JPEG in a wide coding bit-rate range.

\item The proposed method can be easily extended to other compression standards, such as JPEG 2000, H.264, HEVC, etc. In addition, this work provides some insights on the SR of low quality LR images (e.g., noisy and blurry), and more attention would be attracted to concern this kind of problems.
\end{itemize}

The rest of this paper is organized as follows. Section 2 briefly reviews related works. Section 3 presents the proposed CISRDCNN. Extensive experiments are shown in Section 4. Finally, Section 5 concludes this paper.

\section{Related work}
\subsection{Problem formulation of compressed image SR}
Let ${\bf{X}}$ and ${\bf{Y}}$ be HR and LR images, respectively. Then the conventional SR problem can be formulated as
\begin{equation}
\label{eq.1}
{\bf{Y}} = {\bf{HX}} + n
\end{equation}
where $\bf{H}$ represents the synthetic operator of downsampling and blurring, and $n$ denotes the additive noise. The aim of SR is to obtain a high quality estimation of ${\bf{X}}$ from ${\bf{Y}}$.

In this work, we address the problem of compressed image SR. Therefore, the compression process should be concerned. Let ${\bf{C}}$ be the composite operator of compression and decompression, and thus the Eq. \ref{eq.1} is changed to
\begin{equation}
\label{eq.2}
{\bf{Z}} = {\bf{CHX}}
\end{equation}
where ${\bf{Z}}$ is the compressed LR image. Note that we neglect the additive noise in this work as our focus is compression artifacts. Correspondingly, estimating ${\bf{X}}$ from the compressed LR observation ${\bf{Z}}$ is our goal. For convenience and clarity of representation, we still let ${\bf{Y}} = {\bf{HX}}$, so we have ${\bf{Z}} = {\bf{CY}}$. Hence,  ${\bf{Y}}$ only suffers from blurring and downsampling, but ${\bf{Z}}$ suffers from blurring, downsampling, and compression degradations. The compression process would cause information lost, especially at high compression ratio. Intuitively, it is much harder to recover ${\bf{X}}$ from ${\bf{Z}}$ than from ${\bf{Y}}$. Therefore, we try to use the intermediate observation  ${\bf{Y}}$ to assist the reconstruction process  in the proposed CISRDCNN. Note that the intermediate observation  ${\bf{Y}}$ just exists in training phase, while the only input in testing phase is ${\bf{Z}}$.

Many compression methods have been proposed for still images, nevertheless, JPEG still remains one of the most widely used standards. Hence, in this work, we take the JPEG as an example to test the performance of the proposed CISRDCNN.

\subsection{Deep neural networks for image SR}
Deep neural networks have been widely used to address image restoration problems, including SR \cite{37,38,39,40,41,42,43,52,53}, denoising \cite{53,52}, CAR \cite{52,53,54,55}, deblurring \cite{56}, dehazing \cite{57}, etc. In this section, we review some relevant deep neural networks-based SR methods.

In \cite{37}, Dong et al. proposed a CNN-based SR framework (SRCNN), which is composed of three convolutional layers. The three layers realize patch extraction, non-linear mapping, and reconstruction, respectively. The SRCNN has drawn wide attention for its excellent performance and simple network architecture. Later on, Dong et al. \cite{40} presented an accelerated version of SRCNN, which is named FSRCNN. The FSRCNN incorporates the upsampling operation into the network and has a hourglass-shape structure, thus achieving remarkable restoration quality and fast execution speed. By contrast, the SRCNN and FSRCNN are relatively shallow. Kim et al. \cite{41} designed a 20-layer CNN (VDSR), which produces great performance enhancement over SRCNN. For more stable training and better performance, residual-learning and gradient clipping are used in VDSR. More recently, Zhang et al. \cite{53} proposed a similar network (DnCNN), which combines more advances on deep learning, including Residual Learning \cite{58}, batch normalization \cite{59}, and Rectifier Linear Unit (ReLU) \cite{60}. The DnCNN shows great effectiveness in several general image denoising problems, including SR and CAR. Overall, the deep neural networks-based SR methods always result in compelling performance, and most of them are efficient in testing phase.

To the best of our knowledge, nevertheless, very few research has been done on deep neural networks-based SR methods for compressed images. The aim of this work is to propose an effective SR method for compressed images using the considerable advances on deep CNN. We try to design a deep network that realizes CAR and SR jointly.

\subsection{Advances on deep neural networks}
In recent years, many advances have been achieved on deep learning. In the following, we introduce some representative achievements related to this work, i.e., residual learning \cite{58}, batch normalization \cite{59}, and ReLU \cite{60}.

\subsubsection{Residual learning}
In \cite{58}, He et al. firstly proposed the residual learning strategy to address the performance degradation problem caused by the increase of network depth. The main assumption of residual learning is that the learning of residual mapping is much easier than the original mapping. With the residual learning framework, deeper network can be designed and well trained, thus achieving better performance. Actually, similar idea has also been used in many learning-based SR methods, in which the residual image between ground truth and initialization (generally, the interpolated image) is predicted, e.g., ScSR \cite{29}, ANR \cite{33}, and A+ \cite{34}.

\subsubsection{Batch normalization}
In order to ease the internal covariate shift, Ioffe et al. \cite{59} presented the batch normalization. Internal covariate shift refers to the change in the distribution of each layer's input, which slows down the training process and makes it harder. To address this problem, Ioffe et al. proposed to normalize layer inputs. More specifically, in each layer, a normalization step and a scale and shift step are incorporated before the nonlinearity. To realize batch normalization, two extra parameters are added per activation, and these parameters can be learned in network training stage. The batch normalization brings a lot of benefits, such as strong robustness to initialization, fewer training steps, and better performance.

\subsubsection{ReLU}
ReLU is a commonly used activation function in deep neural networks, which outputs $0$ for non-positive inputs and retains positive inputs \cite{60}. The definition of ReLU is:

\begin{equation}
\label{eq.3}
f(x) = \left\{ \begin{array}{l}
x{\kern 1pt} {\kern 1pt} {\kern 1pt} {\kern 1pt} {\kern 1pt} {\kern 1pt} {\kern 1pt} {\kern 1pt} {\kern 1pt} if{\kern 1pt} {\kern 1pt} {\kern 1pt} x > 0\\
0{\kern 1pt} {\kern 1pt} {\kern 1pt} {\kern 1pt} {\kern 1pt} {\kern 1pt} {\kern 1pt} {\kern 1pt} {\kern 1pt} else
\end{array} \right.
\end{equation}

The ReLU alleviates the gradient vanishing problem to some extent, thus making the training of deep neural networks easier.

For fast and stable training procedure and excellent restoration quality, the proposed CISRDCNN integrates residual learning, batch normalization, and ReLU. Unlike the DnCNN \cite{53} that employs a single residual unit, the CISRDCNN uses two residual units due to the fact that the input and output of CISRDCNN are different in resolution. More details about CISRDCNN will be introduced in Section 3.

\section{The proposed CISRDCNN}

In this section, we present the CISRDCNN in detail. As illustrated in Fig. \ref{figure.2}, the proposed CISRDCNN consists of three modules: deblocking module (DBCNN), upsampling module (USCNN), and quality enhancement module (QECNN). Firstly, the DBCNN removes compression artifacts in input and generates a better input for USCNN. Secondly, the USCNN magnifies its input to expected resolution, and thus no extra interpolation procedure is needed. Finally, the QECNN is integrated to improve the quality of upsampled image further. Although the three modules have their respective functions, they are not independent and the whole network is end-to-end trainable. Overall, the differences between CISRDCNN and relevant CNN-based SR methods (e.g., SRCNN \cite{37}, FSRCNN \cite{40}, VDSR \cite{41}, DnCNN \cite{53}) are mainly in the following aspects:

\begin{itemize}

\item The CISRDCNN is for compressed images, and thus compression noise is taken into consideration carefully. Nevertheless, most of the learning-based SR methods do not apply to noisy images.

\item The CISRDCNN is composed of three functional modules, however, it is end-to-end trainable. In this way, the functions of the three modules can be preserved to some extent; meanwhile, the whole network can be optimized to produce minimum prediction error.

\item The CISRDCNN is trained in a particular way. It firstly trains the three functional modules separately to achieve their respective goals, then optimizes the whole network jointly with the fine-tuning strategy.

\end{itemize}

In CISRDCNN, these specific design and improvements for compressed images enable more accurate estimation of the ground truth. In the following, more details about the architecture and training strategy of CISRDCNN are presented.

\subsection{Network architecture}

For the convenience of representation, the depths of DBCNN, USCNN, and QECNN are denoted as ${K_1}$, ${K_2}$, and ${K_3}$, respectively.

\subsubsection{DBCNN}
The DBCNN is composed of two types of layers. The first ${K_1}- 1$ convolutional layers use $64$ filters of size $3 \times 3$, and the batch normalization and ReLU are placed behind these convolutional layers as \cite{53}. The last layer (the ${K_1}$-th layer) generates the restored image using $1$ filter of size $3 \times 3$. Since the input and output of DBCNN are very similar, learning the residual image is more suitable. Hence, we adopt residual learning strategy in this module. More specifically, an identity connection is used to pass the input of DBCNN to its output. Note that all of the batch normalization and ReLU are not presented in Fig. \ref{figure.2} for brevity.

\subsubsection{USCNN}
The first ${K_2}- 1$ convolutional layers of USCNN are the same, using $64$ filters of size $3 \times 3$ and followed by batch normalization and ReLU. The last layer is a deconvolutional layer, which performs upsampling operation. The deconvolutional layer produces one upsampled image using $1$ filter of size $9 \times 9$.

\subsubsection{QECNN}
The architecture of QECNN is similar to DBCNN. Therefore, we do not introduce the QECNN in detail to avoid redundancy.
\begin{figure}[!tb]
\centering
\includegraphics{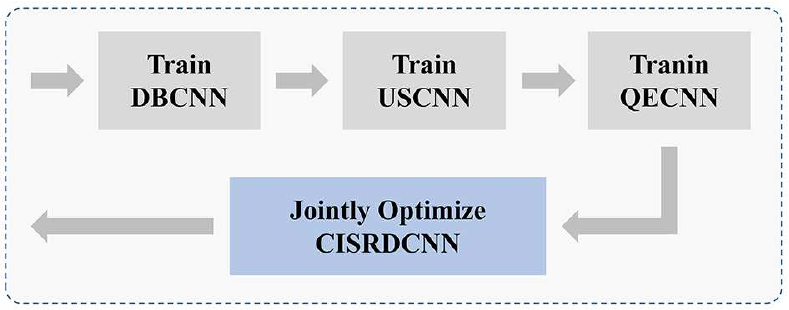}
\captionsetup{justification=centering}
\caption{The flowchart of CISRDCNN training procedure.}
\label{figure.3}
\end{figure}

\subsection{Network training strategy}
Let $\{ {\bf{X}}_i^{train},{\bf{Y}}_i^{train},{\bf{Z}}_i^{train}\} _{i = 1}^N$ be the training image pairs. As introduced in Section \emph{2.1}, ${\bf{X}}_i^{train}$ denotes a HR sample, ${\bf{Y}}_i^{train}$ denotes the corresponding LR sample that only suffers from blurring and downsampling, and ${\bf{Z}}_i^{train}$ represents the compressed version of ${\bf{Y}}_i^{train}$.

As shown in Fig. \ref{figure.3}, the training of CISRDCNN is mainly composed of four steps. Firstly, the set $\{{\bf{Z}}_i^{train},{\bf{Y}}_i^{train}\} _{i = 1}^N$ is used to train the deblocking network DBCNN. As we adopt the residual learning strategy, our goal is to learn a residual mapping ${f_{DB}}({\bf{Z}}^{train})$ that predicts the residual image ${{\bf{r}}_{DB}^{train} = }{\bf{Y}}^{train} - {\bf{Z}}^{train}$. Consequently, the loss function of DBCNN is defined as
\begin{equation}
\label{eq.4}
{l_{DB}}({\Theta _{DB}}) = \frac{1}{{2N}}\sum\limits_{i = 1}^N {{{\left\| {{f_{DB}}({\bf{Z}}_i^{train};{\Theta _{DB}}) - {\bf{r}}_{DB,i}^{train}} \right\|}_F^2}}
\end{equation}
where ${\Theta _{DB}}$ denotes the trainable parameter set in DBCNN, and ${\bf{r}}_{DB,i}^{train} = {\bf{Y}}_i^{train} - {\bf{Z}}_i^{train}$.

Secondly, we train the upsampling network USCNN. Once the training of DBCNN is finished, we can get the estimation of ${\bf{Y}}_i^{train}$ (denoted as ${\bf{\hat Y}}_i^{train}$) from its compressed observation ${\bf{Z}}_i^{train}$. The training set for USCNN is $\{{\bf{\hat Y}}_i^{train},{\bf{X}}_i^{train}\} _{i = 1}^N$. That is, USCNN aims to learn a function ${f_{US}}({{\bf{\hat Y}}}^{train})$ that maps  ${\bf{\hat Y}}^{train}$ to ${\bf{X}}^{train}$. Formally, the loss function of USCNN is defined as
\begin{equation}
\label{eq.5}
{l_{US}}({\Theta _{US}}) = \frac{1}{{2N}}\sum\limits_{i = 1}^N {\left\| {{f_{US}}({\bf{\hat Y}}_i^{train};{\Theta _{US}}) - {\bf{X}}_i^{train}} \right\|_F^2}
\end{equation}
where ${\Theta _{US}}$ denotes the trainable parameter set in USCNN.

Thirdly, we train the quality enhancement network QECNN. Similarly, the HR version of ${\bf{Z}}^{train}$ is estimated using the learned DBCNN and USCNN, and the estimation is denoted as ${\bf{\hat X}}^{train}$. Correspondingly, the training set for QECNN is $\{{\bf{\hat X}}_i^{train},{\bf{X}}_i^{train}\} _{i = 1}^N$. In QECNN, we also adopt the residual learning, and thus the goal is to learn  a residual mapping ${f_{QE}}({\bf{\hat X}}^{train})$ that predicts the residual image ${{\bf{r}}_{QE}^{train} = }{\bf{X}}^{train} - {\bf{\hat X}}^{train}$. Hence, the loss function of QECNN is defined as
\begin{equation}
\label{eq.6}
{l_{QE}}({\Theta _{QE}}) = \frac{1}{{2N}}\sum\limits_{i = 1}^N {{{\left\| {{f_{QE}}({\bf{\hat X}}_i^{train};{\Theta _{QE}}) - {\bf{r}}_{QE,i}^{train}} \right\|}_F^2}}
\end{equation}
where ${\Theta _{QE}}$ denotes the trainable parameter set in QECNN, and ${\bf{r}}_{QE,i}^{train} = {\bf{X}}_i^{train} - {\bf{\hat X}}_i^{train}$.

Finally, the CISRDCNN is optimized in an end-to-end manner. The learned parameters of DBCNN, USCNN, and QECNN are used to initialize CISRDCNN firstly, and then we use the training sample set $\{{\bf{Z}}_i^{train},{\bf{X}}_i^{train}\} _{i = 1}^N$ to optimize the whole network with the fine-tuning strategy. The loss function for the end-to-end optimization procedure is defined as
\begin{equation}
\label{eq.7}
{l_{CI}}({\Theta _{CI}}) = \frac{1}{{2N}}\sum\limits_{i = 1}^N {\left\| {{f_{CI}}({\bf{Z}}_i^{train};{\Theta _{CI}}) - {\bf{X}}_i^{train}} \right\|_F^2}
\end{equation}
where ${\Theta _{CI}}$ denotes the trainable parameter set in CISRDCNN.

In CISRDCNN, the three modules are with specific functions, i.e., deblocking, upsampling, and quality enhancement. With the above training strategy, the goal of each module can be achieved, while the final joint optimization procedure minimizes prediction error. On the other hand, training a deep network directly is hard. Initializing the deep network with learned parameters is beneficial to obtaining stable training procedure and fast convergence speed. Note that for different compression quality factors (QFs), the networks can also be trained from the learned model with the fine-tuning strategy, rather than training from scratch.

\begin{figure}[!tb]
\centering
\includegraphics[width=10cm]{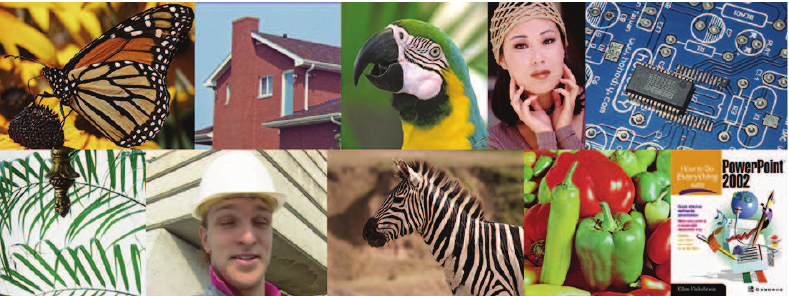}
\caption{Test images (Set10), from left to right and top to bottom: \emph{Butterfly} ($639\times480$), \emph{House} ($256\times256$), \emph{Parrot} ($256\times256$), \emph{Woman} ($321\times481$),  \emph{Circuit} ($694\times502$), \emph{Leaves} ($256\times256$), \emph{Foreman} ($336\times272$), \emph{Zebra} ($700\times523$), \emph{Peppers} ($496\times496$), \emph{Ppt3} ($529\times656$).}
\label{figure.4}
\end{figure}

\begin{table*}[!t]

\scriptsize
\centering
\renewcommand{\arraystretch}{1.55}
\caption{\newline PSNR (\textnormal{d}B) scores of different methods on S\textnormal{et}10 (SR factor: 2, QF: 10/20/30).}

\resizebox{\textwidth}{!}{
 \begin{tabular}{p{4cm}<{\centering}p{1cm}<{\centering}p{1cm}<{\centering}p{1cm}<{\centering}p{1cm}<{\centering}p{1cm}<{\centering}p{1cm}<{\centering}
 p{1cm}<{\centering}p{1cm}<{\centering}p{1cm}<{\centering}p{1cm}<{\centering}p{1cm}<{\centering}}
\hline
\emph{Test Images}  &  \emph{Butterfly}  &  \emph{House} &  \emph{Parrot}  & \emph{Woman} &  \emph{Circuit}  &  \emph{Leaves}  & \emph{Foreman}  & \emph{Zebra}  & \emph{Peppers} & \emph{Ppt3} & Average\\
\hline
\multicolumn{12}{c}{Quality Factor = 10}\\
\hline
Bicubic & 22.691  & 27.392  & 26.234  & 26.284  & 23.373   & 21.455  & 27.759 & 25.654  &  28.489    &  23.258  & 25.259   \\

 A+ \cite{34} & 23.182 & 27.915  & 26.642  & 27.046  & 23.832   & 21.905  & 28.801 & 26.375   & 29.333  & 23.159   & 25.819   \\

FSRCNN \cite{40} & 23.908 &  28.676 & 27.214  & 27.522  &  24.636  &  22.974 &  29.302   & 26.784  & 29.809  & 24.577   & 26.540   \\

VDSR \cite{41} & 24.193  & 29.439  & 26.502  & 27.873  &  24.444  & 22.676  & 29.735   & 26.838  & 30.144  &  24.749  &  26.659  \\

CONCOLOR-VDSR \cite{61,41} & 23.454 & 28.292  & 26.918  & 27.323  &  24.397    & 22.336 & 29.555  & 26.801  & 29.795  &  24.202  & 26.307   \\

ARCNN-VDSR \cite{54,41} & 24.115 & 28.388  &  27.153 & 27.526  & 24.586   &  23.222   & 29.602  &  26.668 & 29.846  & 24.748   & 26.585   \\

SRCDFOE \cite{46} & 23.174 & 28.093  & 26.653  &  26.876 &  23.896  & 21.840  &  28.374   &  26.271 & 29.351  &  23.722  &  25.825  \\

LJSRDB \cite{49} & 22.667 &  27.591 & 26.329  & 26.336  &  23.393  &   21.451 &  28.126 & 25.765  &  28.798 &  23.198  &  25.365  \\

Proposed CISRDCNN  & \textbf{24.534} &  \textbf{30.034} &  \textbf{27.545} &  \textbf{28.177} & \textbf{25.131}   &  \textbf{23.761}  &  \textbf{30.271} & \textbf{27.335}  &  \textbf{30.381} &   \textbf{25.188} & \textbf{27.236} \\

\hline
\multicolumn{12}{c}{Quality Factor = 20}\\
\hline
Bicubic & 23.888 & 29.122  & 27.637  & 27.923  &  25.000  & 23.074  & 29.740 &  27.243  & 30.452  & 24.511   &  26.859  \\

 A+ \cite{34} & 24.570 &  29.888 & 27.971  & 28.796  &  25.708  &  23.894  &  31.072 & 28.192  & 31.411  & 24.891   & 27.639   \\

FSRCNN \cite{40} & 25.374 &  30.528 & 28.589  & 29.219  &  26.433  & 24.907  &  31.642   & 28.479  &  31.705 &  26.203  & 28.308   \\

VDSR \cite{41} & 25.491 & 31.292  & 28.164  &  29.450 & 26.167   & 24.844  &  31.865  & 28.366  &  31.966 &  26.387  &  28.399  \\

CONCOLOR-VDSR \cite{61,41} & 24.628 & 29.344  & 28.093  &  28.804 & 25.779   &  23.969  & 31.683  & 28.109  &  31.415 & 25.941   &  27.777  \\

ARCNN-VDSR \cite{54,41} & 25.429 &  29.536 &  28.531 & 28.990  &  26.235  &  24.951   &  31.647 & 28.241  & 31.652  &  26.470  & 28.168   \\

SRCDFOE \cite{46} & 24.423 &  29.809 &  27.976 & 28.509  &  25.515  & 23.588  &   30.343 &  27.819 &  31.141 &  25.194  &  27.432  \\

LJSRDB \cite{49} & 23.858 & 28.991  & 27.615  &  27.832 &  24.686  &  22.777  &  29.922 &  27.148 & 30.427  &  24.455  & 26.771   \\

Proposed CISRDCNN & \textbf{25.931} &  \textbf{31.652} &  \textbf{28.983} &  \textbf{30.026} & \textbf{26.947}   &  \textbf{26.102}  &  \textbf{32.246} & \textbf{28.951}  &  \textbf{32.194} &   \textbf{27.179} & \textbf{29.021}\\
\hline
\multicolumn{12}{c}{Quality Factor = 30}\\
\hline
Bicubic & 24.508 &  29.660 & 28.510  &  28.807 &  25.738  & 23.962  & 30.724   & 28.132  &  31.355 &  25.130  & 27.653   \\

 A+ \cite{34} & 25.289 & 30.431  &  29.066 & 29.742  &  26.568  & 24.980   &  32.194 & 29.162  &  32.342 & 25.838   & 28.561   \\

FSRCNN \cite{40} & 26.089 &  31.103 &  29.554 & 30.198  &   27.310  & 26.148 & 32.610  & 29.326  & 32.594  &  27.141  & 29.207   \\

VDSR \cite{41} & 26.304 & 31.878  &  29.354 & 30.500  &   27.293 & 26.145 & 32.991  & 29.465  & 32.860  & 27.752   &  29.454  \\

CONCOLOR-VDSR \cite{61,41} & 25.316 &  29.992 & 29.085  &  29.630 & 26.455   &  24.972  & 32.405  & 28.975  & 32.252  &  27.091  &  28.617  \\

ARCNN-VDSR \cite{54,41} & 26.257 &  29.816 & 29.518  & 30.221  &  27.048  & 26.080  &  32.586  &  29.376 & 32.663  &   27.637 &  29.120  \\

SRCDFOE \cite{46} &  25.108 & 30.351  & 28.926  & 29.413  &  26.248  & 24.601  & 31.275 & 28.792  &  31.982 &  26.050  & 28.275   \\

LJSRDB \cite{49} & 24.530 & 29.321  &  28.548 &  28.752 &  24.853  & 23.286  &  30.621  & 27.987  & 31.238  &  24.843  &  27.398  \\

Proposed CISRDCNN & \textbf{26.646} &  \textbf{32.214} &  \textbf{29.890} &  \textbf{30.918} & \textbf{27.847}   &  \textbf{27.305}  &  \textbf{33.257} & \textbf{29.922}  &  \textbf{33.074} &   \textbf{28.481} & \textbf{29.955} \\
\hline
\end{tabular}}
\label{table.1}
\end{table*}

\begin{table*}[!t]
\centering
\scriptsize
\renewcommand{\arraystretch}{1.55}
\caption{\newline SSIM scores of different methods on S\textnormal{et}10 (SR factor: 2, QF: 10/20/30).}
\centering
\resizebox{\textwidth}{!}{
 \begin{tabular}{p{4cm}<{\centering}p{1cm}<{\centering}p{1cm}<{\centering}p{1cm}<{\centering}p{1cm}<{\centering}p{1cm}<{\centering}p{1cm}<{\centering}
 p{1cm}<{\centering}p{1cm}<{\centering}p{1cm}<{\centering}p{1cm}<{\centering}p{1cm}<{\centering}}
\hline
\emph{Test Images}  &  \emph{Butterfly}  &  \emph{House} &  \emph{Parrot}  & \emph{Woman} &  \emph{Circuit}  &  \emph{Leaves}  & \emph{Foreman}  & \emph{Zebra}  & \emph{Peppers} & \emph{Ppt3} & Average\\
\hline
\multicolumn{12}{c}{Quality Factor = 10}\\
\hline
Bicubic & 0.6790  & 0.7499  &  0.7742 &  0.7684 & 0.6956   & 0.6680  & 0.7771 & 0.7996 & 0.7550    & 0.7876   &  0.7454  \\

 A+ \cite{34} & 0.7180 & 0.7870  & 0.8054  & 0.8098  & 0.7317   & 0.7172  &0.8224  & 0.8342   & 0.7914  & 0.8231   & 0.7840   \\

FSRCNN \cite{40} & 0.7437 & 0.8084  &0.8250   & 0.8319  & 0.7694   & 0.7953  & 0.8450    & 0.8513  & 0.8081  & 0.8721   & 0.8150   \\

VDSR \cite{41} & 0.7575&0.8232   & 0.8279  & 0.8406  & 0.7741   & 0.7976  & 0.8551   & 0.8562  & 0.8150  &0.8831    &  0.8230  \\

CONCOLOR-VDSR \cite{61,41} & 0.7422  & 0.8057  &  0.8255 & 0.8317  & 0.7705     & 0.7549 & 0.8486  & 0.8556  & 0.8118  & 0.8697   &  0.8116  \\

ARCNN-VDSR \cite{54,41} & 0.7494 &  0.8050 & 0.8278  & 0.8317 & 0.7678   &  0.8023   & 0.8477  & 0.8466  &  0.8089 & 0.8720   &  0.8159  \\

SRCDFOE \cite{46} & 0.7213 &  0.7936 &  0.8102 & 0.8113  & 0.7404   &  0.7164& 0.8191    & 0.8369  & 0.7971  & 0.8451   &  0.7891   \\

LJSRDB \cite{49} & 0.6888 & 0.7706  & 0.7921  & 0.7827  & 0.7029   & 0.6728   & 0.8013  &  0.8204 &  0.7781 & 0.8039   &  0.7614  \\

Proposed CISRDCNN  & \textbf{0.7699} &  \textbf{0.8310} &  \textbf{0.8410} &  \textbf{0.8494} & \textbf{0.7937}   &  \textbf{0.8326}  &  \textbf{0.8640} & \textbf{0.8655}  &  \textbf{0.8222} &   \textbf{0.8958} & \textbf{0.8365} \\

\hline
\multicolumn{12}{c}{Quality Factor = 20}\\
\hline
Bicubic & 0.7369 & 0.7999  & 0.8211  & 0.8178  & 0.7596   & 0.7404& 0.8273 &  0.8479  & 0.8029  & 0.8299   &  0.7984  \\

 A+ \cite{34} & 0.7704 & 0.8263  & 0.8414  & 0.8500  & 0.7996   &  0.8059  & 0.8652  & 0.8718  & 0.8283  &0.8748    &  0.8334  \\

FSRCNN \cite{40} & 0.7916 & 0.8410  & 0.8547  & 0.8666  & 0.8234   & 0.8570  & 0.8818    & 0.8819  & 0.8380  & 0.9056   &  0.8542  \\

VDSR \cite{41} & 0.8014 & 0.8478  & 0.8576  & 0.8735  & 0.8288   & 0.8676  & 0.8868   & 0.8856  & 0.8431  & 0.9202    &  0.8612  \\

CONCOLOR-VDSR \cite{61,41} &0.7835  & 0.8301  & 0.8535  & 0.8634  & 0.8159   & 0.8166   & 0.8863  & 0.8815  & 0.8369  & 0.9036   & 0.8471   \\

ARCNN-VDSR \cite{54,41} & 0.7902 & 0.8370  & 0.8548  & 0.8616  & 0.8186   &  0.8550   & 0.8795  & 0.8762  & 0.8365  & 0.9046   &  0.8514  \\

SRCDFOE \cite{46} & 0.7694 & 0.8317  & 0.8427  & 0.8488  & 0.7980  & 0.7939  & 0.8582   & 0.8701  & 0.8280  & 0.8820   &  0.8323  \\

LJSRDB \cite{49} & 0.7444 & 0.8164  &  0.8321 &  0.8265 &  0.7638  & 0.7409   & 0.8446  & 0.8562  &  0.8141 & 0.8433   & 0.8082   \\

Proposed CISRDCNN  & \textbf{0.8135} &  \textbf{0.8533} &  \textbf{0.8678} &  \textbf{0.8822} & \textbf{0.8429}   &  \textbf{0.8949}  &  \textbf{0.8937} & \textbf{0.8938}  &  \textbf{0.8476} &   \textbf{0.9356} & \textbf{0.8725} \\
\hline
\multicolumn{12}{c}{Quality Factor = 30}\\
\hline
Bicubic &  0.7656 & 0.8200  & 0.8440  & 0.8440  & 0.7889   & 0.7792  &  0.8527  & 0.8698  &  0.8225 & 0.8539   &  0.8241  \\

 A+ \cite{34} & 0.7966 & 0.8416  & 0.8638  & 0.8715  & 0.8263   & 0.8441   & 0.8847  & 0.8896  & 0.8431  & 0.8973   &  0.8559  \\

FSRCNN \cite{40} & 0.8135 & 0.8513  & 0.8749  & 0.8846  & 0.8468    & 0.8845 & 0.8953  & 0.8966  & 0.8501  &  0.9232  &  0.8721  \\

VDSR \cite{41} & 0.8243 &  0.8574 &  0.8798 & 0.8929  & 0.8544   & 0.8965 & 0.9021  & 0.9023  & 0.8556  & 0.9404   &  0.8806  \\

CONCOLOR-VDSR \cite{61,41} & 0.8047 & 0.8507  & 0.8734  & 0.8823  & 0.8359   & 0.8505   & 0.8979  & 0.8960  & 0.8490  & 0.9257   & 0.8666   \\

ARCNN-VDSR \cite{54,41} & 0.8165 & 0.8509  & 0.8748  & 0.8853  & 0.8412   & 0.8840  & 0.8948   & 0.8961  & 0.8504  & 0.9249   &  0.8719  \\

SRCDFOE \cite{46} & 0.7946 & 0.8428  & 0.8625  & 0.8688  & 0.8216   & 0.8318  & 0.8756 & 0.8879  & 0.8411  & 0.9034   &  0.8530  \\

LJSRDB \cite{49} & 0.7720 & 0.8269  &  0.8543 & 0.8538  & 0.7842   & 0.7745  & 0.8664   & 0.8738  & 0.8304  & 0.8673   &  0.8304  \\

Proposed CISRDCNN  & \textbf{0.8327} &  \textbf{0.8633} &  \textbf{0.8853} &  \textbf{0.8999} & \textbf{0.8638}   &  \textbf{0.9158}  &  \textbf{0.9081} & \textbf{0.9082}  &  \textbf{0.8588} &   \textbf{0.9515} & \textbf{0.8887}   \\
\hline
\end{tabular}}
\label{table.2}
\end{table*}

\begin{table*}[!t]
\centering
\scriptsize
\renewcommand{\arraystretch}{1.5}
\caption{\newline IFC scores of different methods on S\textnormal{et}10 (SR factor: 2, QF: 10/20/30).}
\centering
\resizebox{\textwidth}{!}{
 \begin{tabular}{p{4cm}<{\centering}p{1cm}<{\centering}p{1cm}<{\centering}p{1cm}<{\centering}p{1cm}<{\centering}p{1cm}<{\centering}p{1cm}<{\centering}
 p{1cm}<{\centering}p{1cm}<{\centering}p{1cm}<{\centering}p{1cm}<{\centering}p{1cm}<{\centering}}
\hline
\emph{Test Images}  &  \emph{Butterfly}  &  \emph{House} &  \emph{Parrot}  & \emph{Woman} &  \emph{Circuit}  &  \emph{Leaves}  & \emph{Foreman}  & \emph{Zebra}  & \emph{Peppers} & \emph{Ppt3} & Average\\
\hline
\multicolumn{12}{c}{Quality Factor = 10}\\
\hline
Bicubic & 1.404 &  0.867 & 0.858  & 0.983  & 1.671   & 1.828  & 0.847 & 1.133   & 0.898    &  1.260  & 1.175  \\

 A+ \cite{34} & 1.684 & 0.996  & 1.052  & 1.208  & 1.932   & 2.162  & 1.060 & 1.400   & 1.092  & 1.363   & 1.395   \\

FSRCNN \cite{40} & 1.887 & 1.138  & 1.164  & 1.329  & 2.178   & 2.554  & 1.170    &  1.546 & 1.227  & 1.660   & 1.585   \\

VDSR \cite{41} & 2.048 &  1.262 &  1.206 & 1.427  & 2.222   & 2.565  & 1.268   & 1.601  & 1.312  & 1.729   &  1.664  \\

CONCOLOR-VDSR \cite{61,41} & 2.115 & 1.268  & 1.298  & 1.401  &  2.222    & 2.555 & 1.345  & 1.626  &1.310   & 1.716   &  1.686  \\

ARCNN-VDSR \cite{54,41} & 1.982 & 1.183  & 1.211  & 1.347  & 2.183   & 2.676    & 1.229  & 1.540  & 1.247  & 1.696   & 1.629   \\

SRCDFOE \cite{46} & 1.688 & 1.007  & 1.040  & 1.161 &  1.878  & 2.034  &  0.996   & 1.361  &  1.104 & 1.413   &  1.368  \\

LJSRDB \cite{49} & 1.478 & 0.903  & 0.922  & 1.024  &  1.699  & 1.888   &  0.920 & 1.253  & 0.968  & 1.276   &   1.233 \\

Proposed CISRDCNN  & \textbf{2.268} &  \textbf{1.413} &  \textbf{1.395} &  \textbf{1.552} & \textbf{2.464}   &  \textbf{3.070}  &  \textbf{1.430} & \textbf{1.763}  &  \textbf{1.435} &   \textbf{1.901} & \textbf{1.869} \\

\hline
\multicolumn{12}{c}{Quality Factor = 20}\\
\hline
Bicubic & 2.030 &  1.248 & 1.284  & 1.437 &  2.374&  2.572  & 1.298  & 1.640 & 1.355   & 1.719  &    1.696  \\

 A+ \cite{34} & 2.336 & 1.451  &  1.502 & 1.697  & 2.746   &  3.080  & 1.560  &  1.926 &  1.595 & 1.930   & 1.982   \\

FSRCNN \cite{40} & 2.544 & 1.556  & 1.568  & 1.813  & 2.985   & 3.342  & 1.618    &  2.066 & 1.679  & 2.171   & 2.134   \\

VDSR \cite{41} & 2.706 & 1.650  & 1.655  & 1.937  & 3.014   & 3.562  & 1.733   & 2.145  & 1.786  & 2.323   &  2.251  \\

CONCOLOR-VDSR \cite{61,41} & 2.710 & 1.627  & 1.683  &  1.873 &  2.889  & 3.320   & 1.831  &  2.085 & 1.725  & 2.279   & 2.202   \\

ARCNN-VDSR \cite{54,41} & 2.562 & 1.533  & 1.564  &  1.770 & 2.907   &  3.340   & 1.620  & 2.020  &  1.653 & 2.206   &  2.118  \\

SRCDFOE \cite{46} & 2.283 & 1.401  &  1.439 & 1.603  & 2.591   & 2.785  & 1.418   & 1.836  &  1.531 &  1.916  &  1.880  \\

LJSRDB \cite{49} & 2.189 &  1.285 & 1.400  & 1.475  &  2.352  & 2.574   & 1.376  &  1.751 & 1.408  & 1.742   &  1.755  \\

Proposed CISRDCNN & \textbf{2.903} &  \textbf{1.767} &  \textbf{1.804} &  \textbf{2.093} & \textbf{3.276}   &  \textbf{4.047}  &  \textbf{1.855} & \textbf{2.283}  &  \textbf{1.884} &   \textbf{2.529} & \textbf{2.444}   \\
\hline
\multicolumn{12}{c}{Quality Factor = 30}\\
\hline
Bicubic & 2.450 & 1.506  & 1.617  & 1.787  &  2.824  & 3.058  & 1.592   & 1.997  & 1.664  & 2.036   &  2.053  \\

 A+ \cite{34} & 2.751 & 1.728  & 1.853  & 2.046  & 3.196   & 3.609   &  1.864 &  2.276 & 1.912  & 2.296   & 2.353   \\

FSRCNN \cite{40} & 2.942 & 1.805  & 1.915  & 2.165  & 3.452    & 3.877 & 1.894  & 2.402  & 1.985  & 2.542   &  2.498  \\

VDSR \cite{41} & 3.163 &  1.934 & 2.039  & 2.316  & 3.583   & 4.102 & 2.040  & 2.532  & 2.103  & 2.791   &  2.660  \\

CONCOLOR-VDSR \cite{61,41} & 3.096 & 1.905  & 2.023  & 2.201  & 3.297   & 3.810   &  2.042 & 2.400  & 2.003  &  2.686  &  2.546  \\

ARCNN-VDSR \cite{54,41} & 3.006 & 1.815  & 1.917  & 2.171  & 3.363   & 3.885  & 1.893   & 2.400  & 1.993  &  2.628  &  2.507  \\

SRCDFOE \cite{46} & 2.683 & 1.655  & 1.754  & 1.944  & 3.038   &  3.289 & 1.687 & 2.174  & 1.814  &  2.277  &  2.232  \\

LJSRDB \cite{49} & 2.599 & 1.484  & 1.727  & 1.819  &  2.664  & 2.958  & 1.633   & 2.043  & 1.666  & 1.996   & 2.059   \\

Proposed CISRDCNN  & \textbf{3.310} &  \textbf{2.032} &  \textbf{2.155} &  \textbf{2.434} &   \textbf{3.751} & \textbf{4.559}   & \textbf{2.130}  &  \textbf{2.609} &  \textbf{2.189} &   \textbf{2.915} &   \textbf{2.808} \\
\hline
\end{tabular}}
\label{table.3}
\end{table*}

\section{Experimental results}
The experimental settings are introduced firstly, and then extensive results are presented to verify the effectiveness of CISRDCNN in this section, including the test on real low quality web images. In addition, we take the low bit-rate coding as an example to show the application of the proposed CISRDCNN.

\subsection{Experimental settings}
Main parameters of CISRDCNN: in our implementation, we set ${K_1} = 20$, ${K_2} = 10$, and ${K_3} = 10$.

Training data: following \cite{41}, the 291 imageset that consists of 200 images from  BSDS500 \footnote{\label{footnote1}Available: http://www.eecs.berkeley.edu/Research/Projects/CS/vision/\\grouping/segbench .}  and 91 images from Yang et al. \cite{28} is used to train CISRDCNN. To increase the number of samples and improve SR performance, we also adopt data augmentation techniques. To generate LR observations, these HR images are downsampled using the \emph{imresize} function (kernel: \emph{bicubic}, downsampling factor: 2) in Matlab firstly, and then the downsampled images are compressed using JPEG.

Test images: Fig. \ref{figure.4} shows the ten test images (named Set10) used in our experiment, which are widely used to evaluate SR methods in literature. For color images, only the luminance components are processed.

Test datasets: four commonly used datasets in SR problem are used to test the performance of different methods, including Set5 \cite{22}, Set14 \cite{29}, B100 \footref{footnote1}, and Urban100 \cite{26}.

Degradation model: for the simulation experiments, the original HR image is downsampled using the \emph{imresize} function (kernel: \emph{bicubic}, downsampling factor: 2) in Matlab firstly, and then the downsampled image is compressed using JPEG with different QFs.

Comparison baselines: the comparison baselines include Bicubic, A+ \cite{34}, FSRCNN \cite{40}, VDSR \cite{41}, CONCOLOR-VDSR \cite{61,41}, ARCNN-VDSR \cite{54,41}\footnote{ARCNN \cite{54} and CONCOLOR \cite{61} are typical and effective compression artifacts reduction methods.}, SRCDFOE \cite{46}, and LJSRDB \cite{49}. For A+ \cite{34}, FSRCNN \cite{40}, and VDSR \cite{41}, we retrained their models according to our experimental settings. The CONCOLOR-VDSR \cite{61,41} and ARCNN-VDSR \cite{54,41} are cascading methods, which consist of the state-of-the-arts of deblocking and SR methods. The SRCDFOE \cite{46} and LJSRDB \cite{49} are two SR algorithms for JPEG compressed images.

Performance evaluation: resultant images of different methods are evaluated objectively and subjectively. For the simulation experiments, the PSNR, SSIM \cite{62}, and IFC \cite{63} are adopted to perform objective evaluation. For the SR of real world compressed images, we use the no-reference quality metric for SR proposed in \cite{64} to evaluate results objectively.

\subsection{Super-resolution results on synthetic LR images}

\subsubsection{Objective evaluation}

Due to the limited space, we only present the objective scores of different methods at QF = 10/20/30 in this subsection. It can be seen from the results reported in Table~\ref{table.1}, Table~\ref{table.2}, and  Table~\ref{table.3} that the CISRDCNN consistently produces the highest PSNR/SSIM/IFC values. Overall, the VDSR \cite{41} generates the second-best results. The FSRCNN \cite{40} and ARCNN-VDSR \cite{41,54} achieve similar performance, and both of them are slightly inferior to the VDSR \cite{41}. The A+ \cite{34}, CONCOLOR-VDSR \cite{41,61} and SRCDFOE \cite{46} are superior to Bicubic, but the gains are limited to some extent. Compared with Bicubic, the LJSRDB \cite{49} produces worse results in some cases. For A+ \cite{34}, the severe compression noise in LR images causes significant performance degradation as it is sensitive to noise. The SRCDFOE \cite{46} and LJSRDB \cite{49} are unified frameworks for JPEG compressed images, however, they do not handle compression noise well. By contrast, more obvious improvement is produced by the proposed CISRDCNN. For example, at QF = 10, the CISRDCNN achieves average 1.977 dB/0.0911/0.694 PSNR/SSIM/IFC gains over Bicubic, and 0.577 dB/0.0135/0.205 over VDSR \cite{41}. Note that the VDSR \cite{41} is one of the state-of-the-art SR methods. Compared with the SR methods for JPEG compressed images, i.e., the SRCDFOE \cite{46} and LJSRDB \cite{49}, the average PSNR/SSIM/IFC gains are up to 1.411 dB/0.0474/0.501 and 1.871 dB/0.0751/0.636, respectively. Similar results can be observed at QF = 20 and QF = 30. In sum, the CISRDCNN achieves state-of-the-art performance.

\begin{figure*}[!tb] 
\centering
\subfigure[]{
\includegraphics[width=2.5cm]{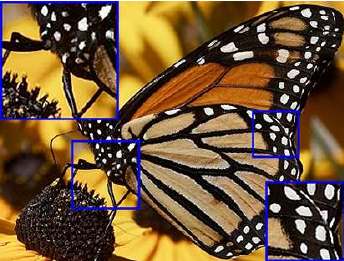}}
\subfigure[]{
\includegraphics[width=2.5cm]{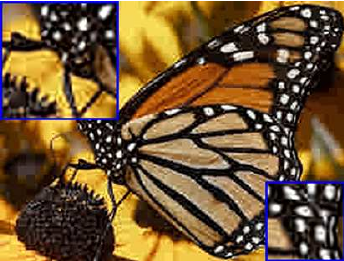}}
\subfigure[]{
\includegraphics[width=2.5cm]{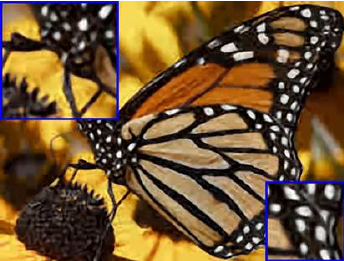}}
\subfigure[]{
\includegraphics[width=2.5cm]{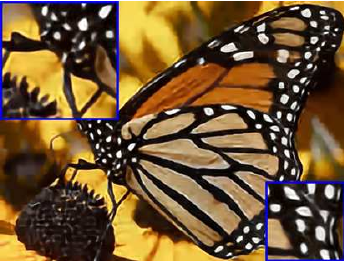}}
\subfigure[]{
\includegraphics[width=2.5cm]{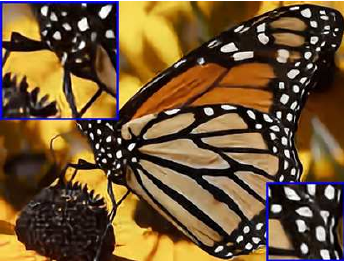}}\\
\subfigure[]{
\includegraphics[width=2.5cm]{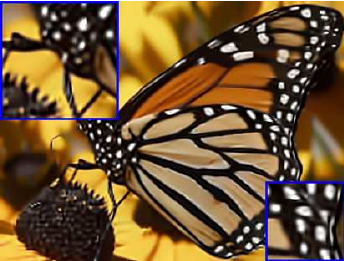}}
\subfigure[]{
\includegraphics[width=2.5cm]{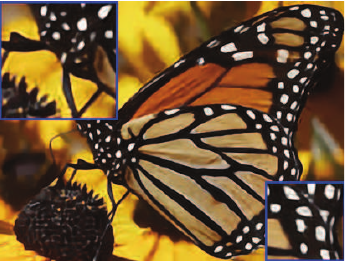}}
\subfigure[]{
\includegraphics[width=2.5cm]{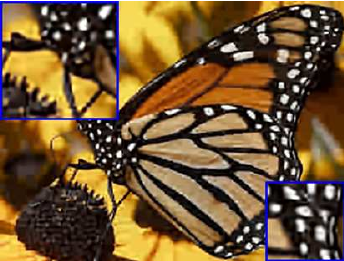}}
\subfigure[]{
\includegraphics[width=2.5cm]{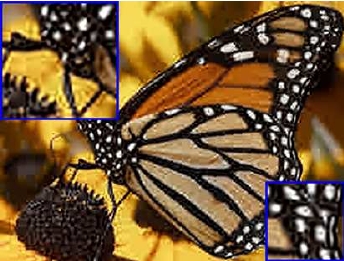}}
\subfigure[]{
\includegraphics[width=2.5cm]{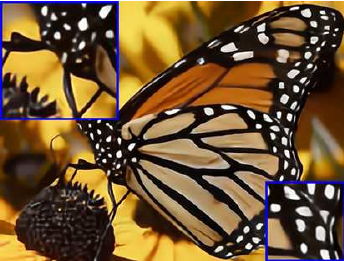}}

\caption{Super-resolution results of \emph{Butterfly} generated by different methods (SR factor: 2, QF: 10). (a) Original image (PSNR (dB), SSIM, IFC). (b) Bicubic (22.691, 0.6790, 1.404). (c) A+ \cite{34} (23.182, 0.7180, 1.684). (d) FSRCNN \cite{40} (23.908, 0.7437, 1.887). (e) VDSR \cite{41} (24.193, 0.7575, 2.048). (f) CONCOLOR-VDSR \cite{61,41} (23.454, 0.7422, 2.115). (g) ARCNN-VDSR \cite{54,41} (24.115, 0.7494, 1.982). (h) SRCDFOE \cite{46} (23.174, 0.7213, 1.688). (i) LJSRDB \cite{49} (22.667, 0.6888, 1.478). (j) Proposed CISRDCNN (\textbf{24.534}, \textbf{0.7699}, \textbf{2.268}). Please zoom in to view details and make comparisons.}
\label{figure.5}
\end{figure*}
\begin{figure*}[!tb] 
\centering
\subfigure[]{
\includegraphics[width=2.5cm]{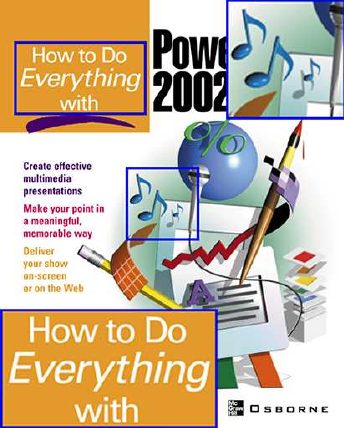}}
\subfigure[]{
\includegraphics[width=2.5cm]{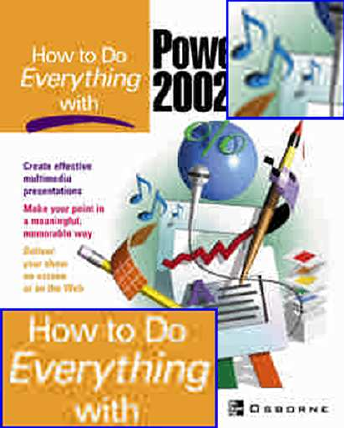}}
\subfigure[]{
\includegraphics[width=2.5cm]{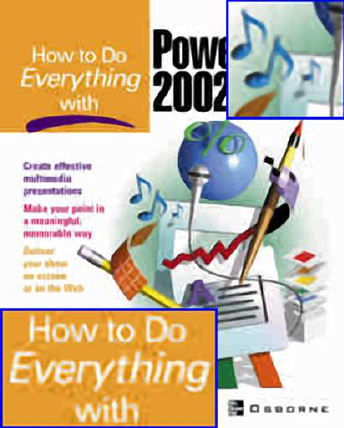}}
\subfigure[]{
\includegraphics[width=2.5cm]{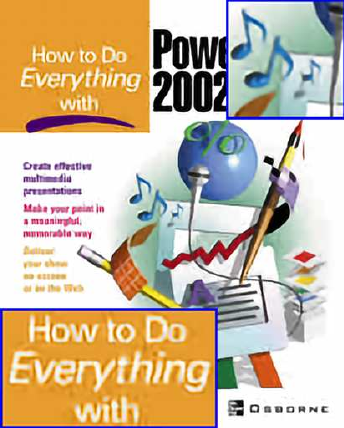}}
\subfigure[]{
\includegraphics[width=2.5cm]{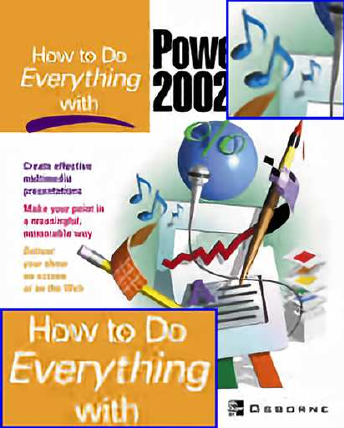}}\\
\subfigure[]{
\includegraphics[width=2.5cm]{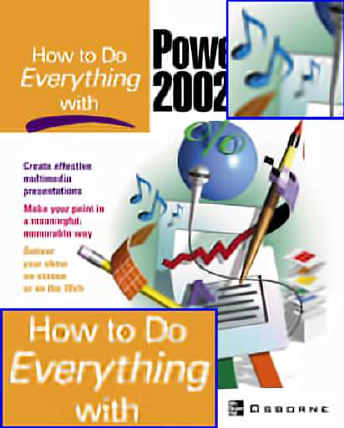}}
\subfigure[]{
\includegraphics[width=2.5cm]{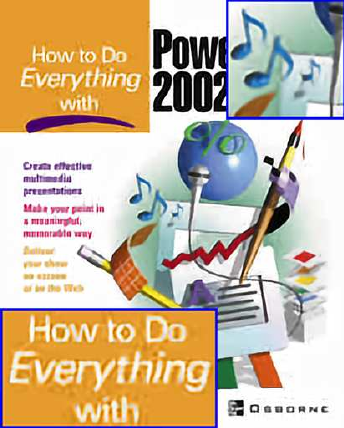}}
\subfigure[]{
\includegraphics[width=2.5cm]{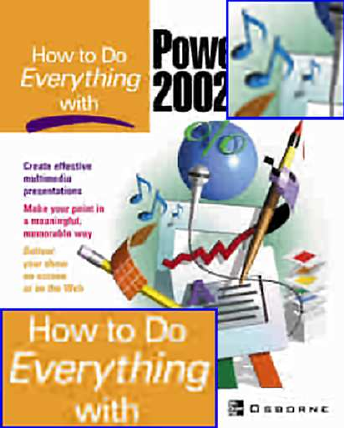}}
\subfigure[]{
\includegraphics[width=2.5cm]{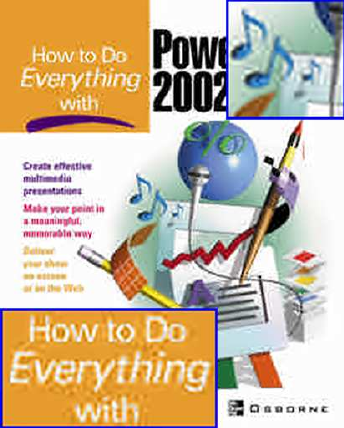}}
\subfigure[]{
\includegraphics[width=2.5cm]{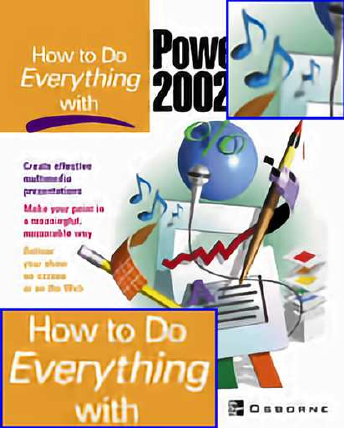}}

\caption{Super-resolution results of \emph{Ppt3} generated by different methods (SR factor: 2, QF: 20). (a) Original image (PSNR (dB), SSIM, IFC). (b) Bicubic (24.511, 0.8299, 1.719). (c) A+ \cite{34} (24.891, 0.8748, 1.930). (d) FSRCNN \cite{40} (26.203, 0.9056, 2.171). (e) VDSR \cite{41} (26.387, 0.9202, 2.323). (f) CONCOLOR-VDSR \cite{61,41} (25.941, 0.9036, 2.279). (g) ARCNN-VDSR \cite{54,41} (26.470, 0.9046, 2.206). (h) SRCDFOE \cite{46} (25.194, 0.8820, 1.916). (i) LJSRDB \cite{49} (24.455, 0.8433, 1.742). (j) Proposed CISRDCNN (\textbf{27.179}, \textbf{0.9356}, \textbf{2.529}). Please zoom in to view details and make comparisons.}
\label{figure.6}
\end{figure*}
\begin{figure*}[!tb] 
\centering
\subfigure[]{
\includegraphics[width=2.5cm]{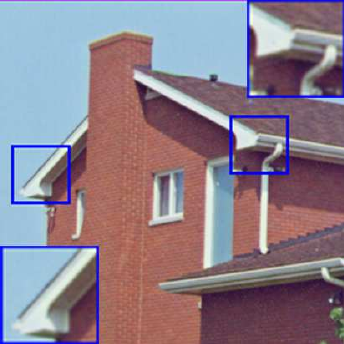}}
\subfigure[]{
\includegraphics[width=2.5cm]{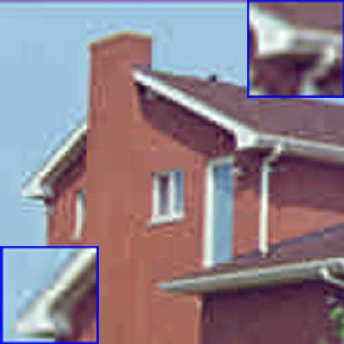}}
\subfigure[]{
\includegraphics[width=2.5cm]{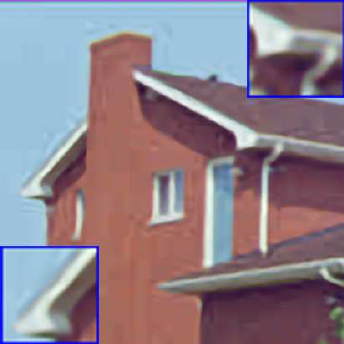}}
\subfigure[]{
\includegraphics[width=2.5cm]{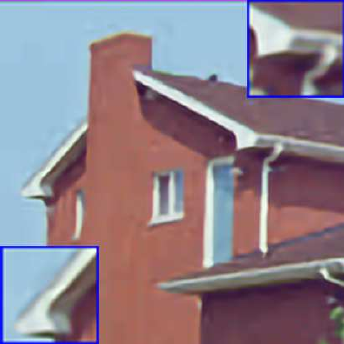}}
\subfigure[]{
\includegraphics[width=2.5cm]{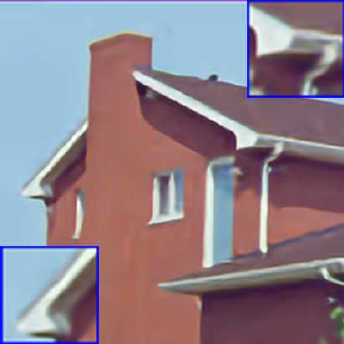}}\\
\subfigure[]{
\includegraphics[width=2.5cm]{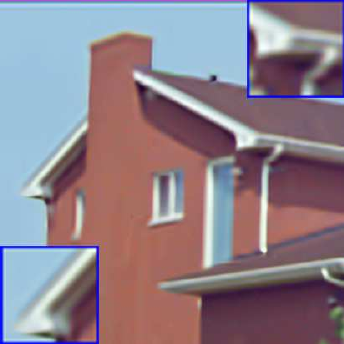}}
\subfigure[]{
\includegraphics[width=2.5cm]{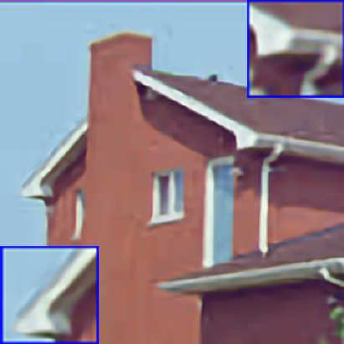}}
\subfigure[]{
\includegraphics[width=2.5cm]{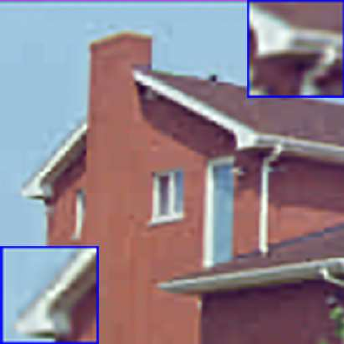}}
\subfigure[]{
\includegraphics[width=2.5cm]{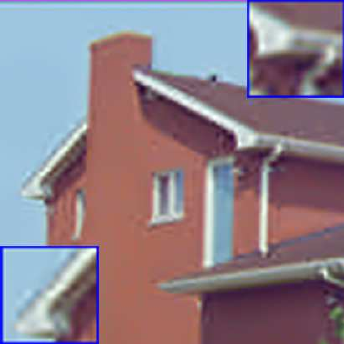}}
\subfigure[]{
\includegraphics[width=2.5cm]{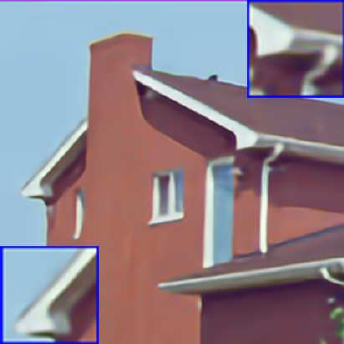}}
\caption{Super-resolution results of \emph{House} generated by different methods (SR factor: 2, QF: 30). (a) Original image (PSNR (dB), SSIM, IFC). (b) Bicubic (29.660, 0.8200, 1.506). (c) A+ \cite{34} (30.431, 0.8416, 1.728). (d) FSRCNN \cite{40} (31.103, 0.8513, 1.805). (e) VDSR \cite{41} (31.878, 0.8574, 1.934). (f) CONCOLOR-VDSR \cite{61,41} (29.992, 0.8507, 1.905). (g) ARCNN-VDSR \cite{54,41} (29.816, 0.8509, 1.815). (h) SRCDFOE \cite{46} (30.351, 0.8428, 1.655). (i) LJSRDB \cite{49} (29.321, 0.8269, 1.484). (j) Proposed CISRDCNN (\textbf{32.214}, \textbf{0.8633}, \textbf{2.032}). Please zoom in to view details and make comparisons.}
\label{figure.7}
\end{figure*}

\subsubsection{Subjective evaluation}

Part of the resultant images are presented to compare visual quality. To comprehensively show the performance of all methods, we deliberately illustrate the results at different QFs. Specifically, Fig. \ref{figure.5} shows the results of \emph{Butterfly} at QF = 10. Fig. \ref{figure.6} shows the results of \emph{Ppt3} at QF = 20. Fig. \ref{figure.7} shows the results of \emph{House} at QF = 30. For better view and comparison, two local regions are highlighted in each figure. The results of Bicubic, A+ \cite{34}, FSRCNN \cite{40}, SRCDFOE \cite{46}, and LJSRDB \cite{49} contain obvious artifacts, especially at low QFs. The VDSR \cite{41}, CONCOLOR-VDSR \cite{61,41}, ARCNN-VDSR \cite{54,41} can remove most of the compression artifacts, nevertheless, the results are blurred somewhat. Comparatively, the results of CISRDCNN are more visually pleasant, with fewer artifacts and clearer structures. For instance, the text in image \emph{Ppt3} (Fig. \ref{figure.6}) and the eave in image \emph{House} (Fig. \ref{figure.7}). In sum, benefitting from the strong ability of deep CNN and the specific design for compressed images, the CISRDCNN realizes joint optimization of compression noise reduction process and SR process, thus leading to state-of-the-art performance.

The results in this subsection provide some insights for the further research on compressed images SR. The comparison set in this experiment is composed of different kinds of methods, including conventional SR method (A+ \cite{34}, FSRCNN \cite{40}, VDSR \cite{41}), cascading SR method (CONCOLOR-VDSR \cite{61,41}, ARCNN-VDSR \cite{54,41}), unified SR framework (SRCDFOE \cite{46}, LJSRDB \cite{49}), and joint optimized SR method (CISRDCNN). According to their performance, we can obtain the following conclusion: the CAR stage is necessary, but it should not be independent of the SR stage. The CAR stage is beneficial to reducing compression artifacts, however, it is hard to control the degree of artifacts reduction. Therefore, joint optimization of CAR and SR is significant. These insights may also apply to the SR of noisy images and blurred images, which will be studied in our future work.

\subsection{Robustness to quality factors}
\begin{figure}[!tb]
\centering
\includegraphics[width=4cm]{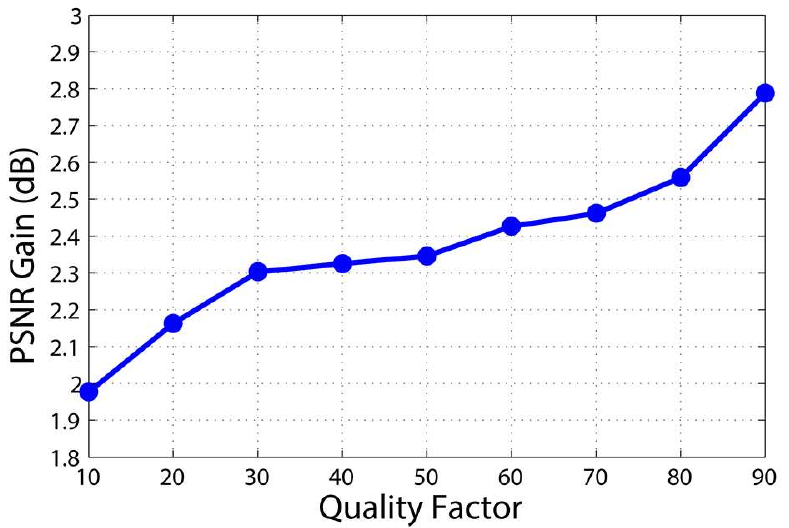}
\caption{The average PSNR gain (dB) of the proposed CISRDCNN over Bicubic at different QFs on Set10.}
\label{figure.8}
\end{figure}

In this subsection, the robustness of CISRDCNN to compression QFs is tested. To conduct this experiment, a series of CNN models are trained at different QFs. Fig. \ref{figure.8} presents the average PSNR gains of CISRDCNN over Bicubic at different QFs on Set10. It can be observed that CISRDCNN achieves obvious PSNR gain in a wide range of QFs, even at low compression ratio. Hence, the CISRDCNN is robust to QFs, and it applies to compressed images in different quality.

\subsection{Experimental results on image datasets}
In order to evaluate the stability and robustness of CISRDCNN on different kinds of images, we conduct experiments on four standard imagesets, including Set5 \cite{22}, Set14 \cite{29}, B100 \footref{footnote1}, and Urban100 \cite{26}. For the images in B100, we crop them to generate test images of size $256 \times 256$. Similarly, the images in Urban100 are cropped to generate small test images of size $512 \times 512$. Due to the limited space, we only take QF = 10 as an example in this experiment, and the Bicubic, VDSR \cite{41}, ARCNN-VDSR \cite{54,41}, and SRCDFOE \cite{46} are selected as baselines. The average PSNR/SSIM/IFC results are reported in Table~\ref{table.4}. It can be observed that the CISRDCNN consistently outperforms all of the compared baselines.

\begin{figure*}[!tb]
\centering
\subfigure[]{
\includegraphics[width=4cm]{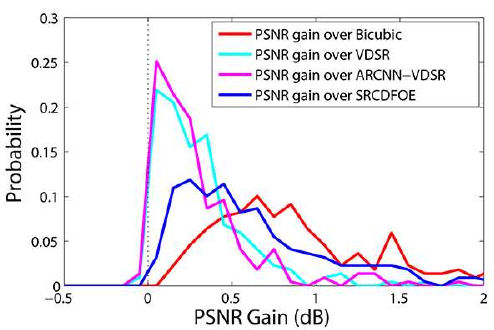}}
\subfigure[]{
\includegraphics[width=4cm]{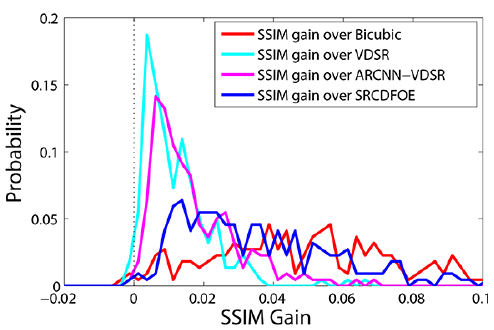}}
\subfigure[]{
\includegraphics[width=4cm]{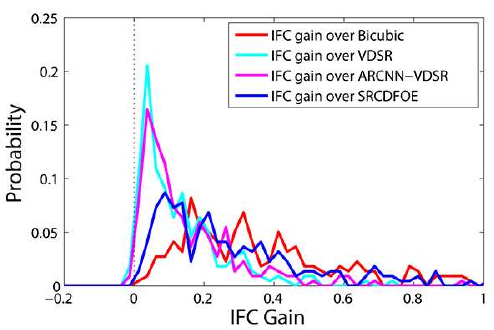}}

\caption{Distributions of PSNR (dB)/SSIM/IFC gains of CISRDCNN over the compared methods, on all images in Set5 \cite{22}, Set14 \cite{29}, B100\footref{footnote1}, and Urban100 \cite{26} (SR factor: 2, QF: 10). (a) Distribution of PSNR (dB) gain. (b) Distribution of SSIM gain. (c)Distribution of IFC gain. The statistical steps for PSNR (dB)/SSIM/IFC gains are set to be 0.1/0.0025/0.025, respectively.  }
\label{figure.9}
\end{figure*}

We further draw the distributions of PSNR/SSIM/IFC gains of CISRDCNN over the baselines in Fig. \ref{figure.9}. One can easily see that the CISRDCNN outperforms competitors for most of the test images in the four commonly used imagesets. The results shown in this subsection demonstrate the robustness and stability of CISRDCNN.

\begin{table}[!tb]
\centering
\tiny
\renewcommand{\arraystretch}{1.5}
\caption{\newline Comparisons of average PSNR (\textnormal{d}B)/SSIM/IFC scores on datasets (SR factor: 2, QF: 10).}
\centering
 \begin{tabular}{p{3.5cm}<{\centering}p{1.5cm}<{\centering}p{1.5cm}<{\centering}p{1.5cm}<{\centering}p{1.5cm}<{\centering}}
\hline
\emph{DataSets}  &  \emph{Set5}  &  \emph{Set14} &  \emph{B100}  & \emph{Urban100} \\
\hline
\multicolumn{5}{c}{PSNR (dB)}\\
\hline
Bicubic & 26.602 & 25.037  & 24.285  &  22.999   \\

VDSR \cite{41} & 27.812 & 25.901  & 24.870  &  23.932   \\

ARCNN-VDSR \cite{54,41} & 27.827 & 25.856  & 24.899  &  23.929    \\

SRCDFOE \cite{46} & 27.243 &  25.451 &  24.599 &  23.340    \\

Proposed CISRDCNN  & \textbf{28.154} &  \textbf{26.127}  & \textbf{25.019}  &  \textbf{24.369}   \\

\hline
\multicolumn{5}{c}{SSIM}\\
\hline
Bicubic & 0.7239 & 0.6433  & 0.5863  &  0.6224    \\

VDSR \cite{41} & 0.7931 & 0.6853  &  0.6179 &   0.6859   \\

ARCNN-VDSR \cite{54,41} & 0.7878 & 0.6803  & 0.6153  &  0.6774    \\

SRCDFOE \cite{46} & 0.7661 &  0.6663 &  0.6025 &   0.6512   \\

Proposed CISRDCNN  & \textbf{0.8039} &  \textbf{0.6926} & \textbf{0.6238}  &  \textbf{0.7043} \\
\hline
\multicolumn{5}{c}{IFC}\\
\hline
Bicubic & 1.036 & 0.989  &  0.817 &  1.142  \\

VDSR \cite{41} & 1.421 &  1.263 &  0.988 & 1.542   \\

ARCNN-VDSR \cite{54,41} & 1.398 & 1.240  & 0.983  &  1.510   \\

SRCDFOE \cite{46} & 1.207 & 1.113  &  0.900 &  1.285   \\

Proposed CISRDCNN  & \textbf{1.546} & \textbf{1.354}  &  \textbf{1.040} &  \textbf{1.751}  \\
\hline
\end{tabular}
\label{table.4}
\end{table}

\subsection{Empirical study on computational time}
\begin{figure*}[!tb]
\centering
\subfigure[]{
\includegraphics[width=4cm]{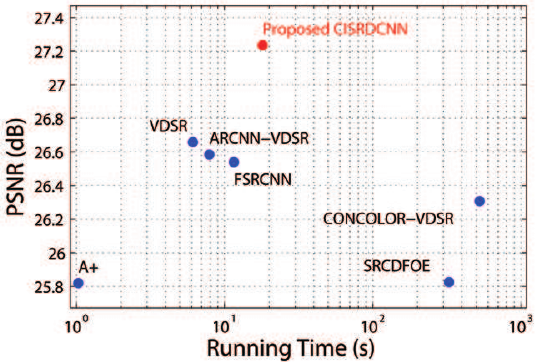}}
\subfigure[]{
\includegraphics[width=4cm]{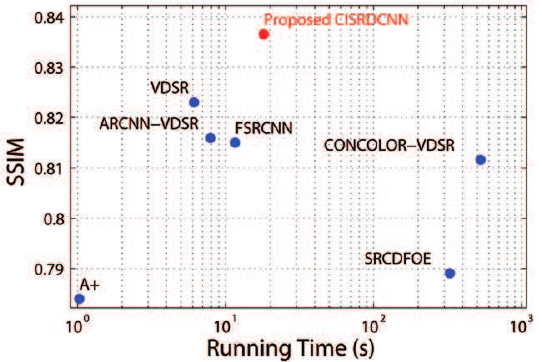}}
\subfigure[]{
\includegraphics[width=4cm]{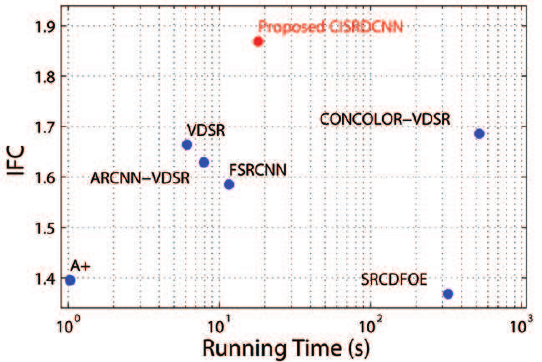}}
\caption{Average PSNR (dB)/SSIM/IFC scores (SR factor: 2, QF: 10) vs. running time (s). (a) PSNR (dB) vs. Running Time (s). (b) SSIM vs. Running Time (s). (c) IFC vs. Running Time (s).}
\label{figure.10}
\end{figure*}

 In this subsection, we compare the running time and PSNR of different methods (QF = 10). This experiment is conducted on a desktop computer (Win7, Inter Core i5 CPU 3.3GHz, 12G memory, Matlab 2014a 64bit) \footnote{The LJSRDB \cite{49} is running on another computer as the code only can run in Matlab 32bit version, so we do not present the running time of this algorithm in Fig. \ref{figure.10}.} . The running time and PSNR of each method are average values of all the ten test images in Fig. \ref{figure.4}. As depicted in Fig. \ref{figure.10}, the proposed CISRDCNN achieves state-of-the-art performance with acceptable computational time\footnote{It is important to note that we use the Matlab test code of FSRCNN (available: http://mmlab.ie.cuhk.edu.hk/projects/FSRCNN.html), which is much slower the implementation used in \cite{40}.}. In addition, the execution time of CISRDCNN can be greatly accelerated with a powerful GPU.

\begin{figure}[!tb] 
\centering
\subfigure[LR (\emph{Rose})]{
\includegraphics[width=3.75cm]{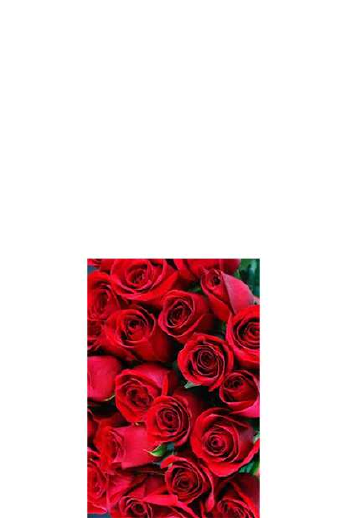}}
\subfigure[Bicubic]{
\includegraphics[width=3.75cm]{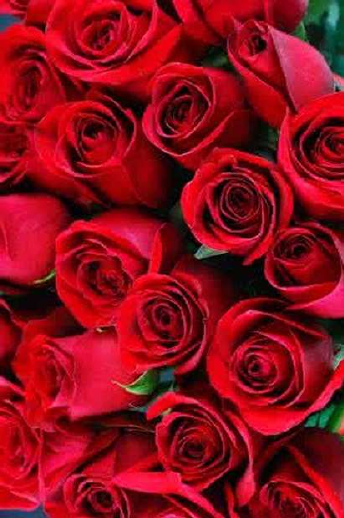}}
\subfigure[CISRDCNN]{
\includegraphics[width=3.75cm]{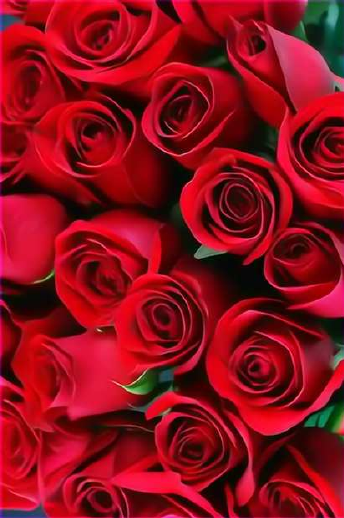}}\\

\subfigure[LR (\emph{Child})]{
\includegraphics[width=3.75cm]{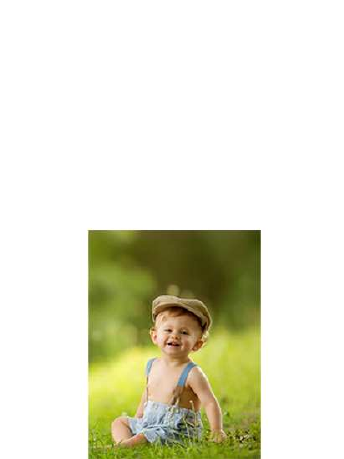}}
\subfigure[Bicubic]{
\includegraphics[width=3.75cm]{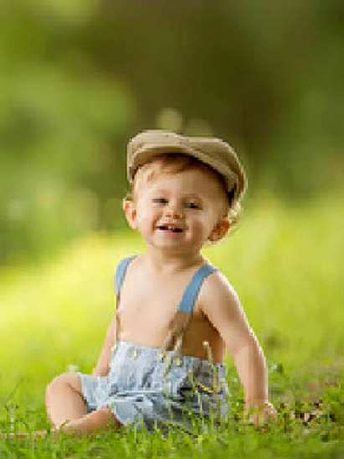}}
\subfigure[CISRDCNN]{
\includegraphics[width=3.75cm]{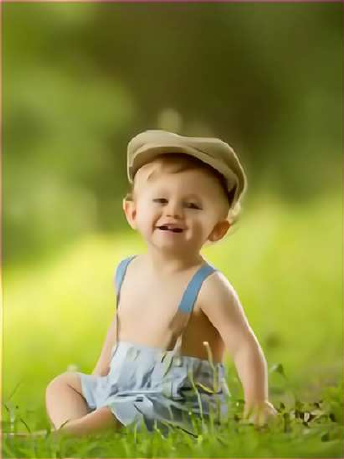}}\\

\subfigure[LR (\emph{Anime})]{
\includegraphics[width=3.75cm]{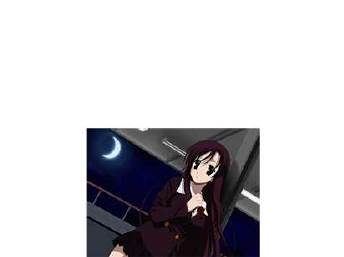}}
\subfigure[Bicubic]{
\includegraphics[width=3.75cm]{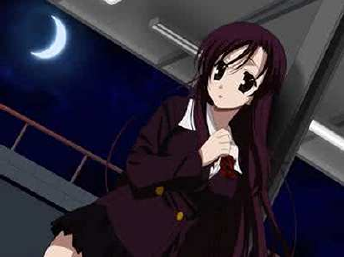}}
\subfigure[CISRDCNN]{
\includegraphics[width=3.75cm]{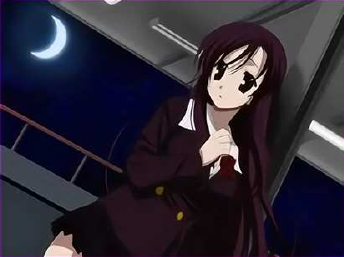}}

\caption{Super-resolution results of real low quality web images (SR factor: 2). The first column [(a)(d)(g)]: real low quality web images. The second column [(b)(e)(h)]: the results of Bicubic. The third column [(c)(f)(i)]: the results of proposed CISRDCNN. Please zoom in to view details and make comparisons.}
\label{figure.11}
\end{figure}

\subsection{Super-resolution on real low quality web images}
We further test the effectiveness of CISRDCNN on real low quality web images, which usually suffer from downsampling and compression due to the limited bandwidth and storage capacity. The test images used in this experiment are downloaded from internet \footnote{Available: http://image.baidu.com .}. As presented in Fig. \ref{figure.11}, we can observe that the CISRDCNN achieves obvious perceptual quality enhancement over the original images and the interpolation results of Bicubic, with fewer artifacts and clearer structures.

Further, the no-reference image quality evaluation index for SR proposed in \cite{64} is used to quantitatively compare these resultant images, and the scores are illustrated in Table \ref{table.5}. It can be seen that the CISRDCNN generates higher values than Bicubic on all of the three test images, which also indicates that the resultant images of CISRDCNN are of higher quality. The results in this subsection verify that the proposed CISRDCNN is also applicable to the compressed image in the real world.

\begin{table}[!tb]
\centering
\tiny

\renewcommand{\arraystretch}{1.5}
\caption{\newline No-reference image quality assessment on the SR results of low quality web images using the evaluation metric for SR proposed in \cite{64} (SR factor: 2).}
 \begin{tabular}{p{2.5cm}<{\centering}p{1cm}<{\centering}p{1cm}<{\centering}p{1cm}<{\centering}}
\hline
\emph{Test Images}  &  \emph{Rose}   &  \emph{Child}  & \emph{Anime} \\
\hline
Bicubic & 2.927 & 3.650 & 3.044   \\

CISRDCNN  & \textbf{4.941} & \textbf{3.902} & \textbf{6.050}   \\
\hline
\end{tabular}
\label{table.5}
\end{table}

\begin{figure}[!tb]
\centering
\includegraphics{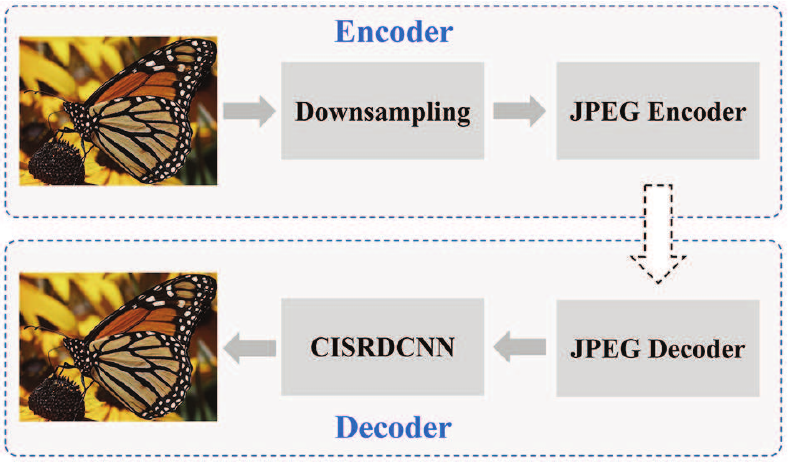}
\captionsetup{justification=centering}
\caption{The flowchart of CISRDCNN-based low bit-rate coding method.}
\label{figure.12}
\end{figure}

\begin{figure*}[!tb]
\centering
\subfigure[]{
\includegraphics[width=4cm]{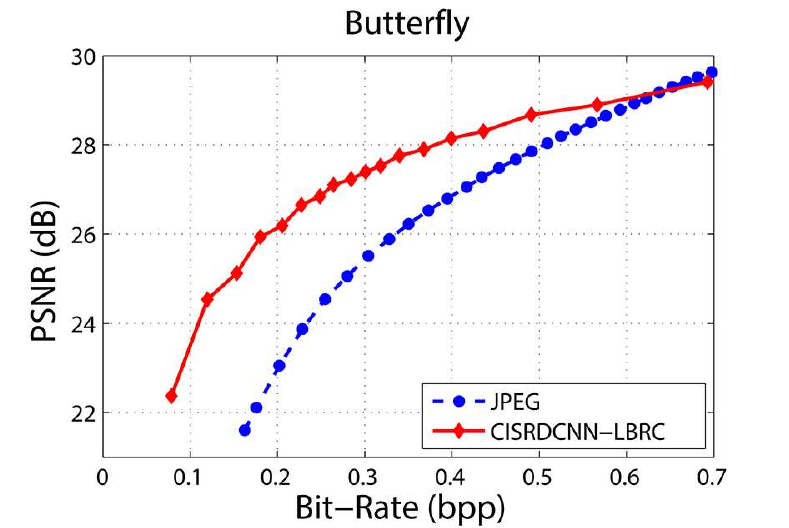}}
\subfigure[]{
\includegraphics[width=4cm]{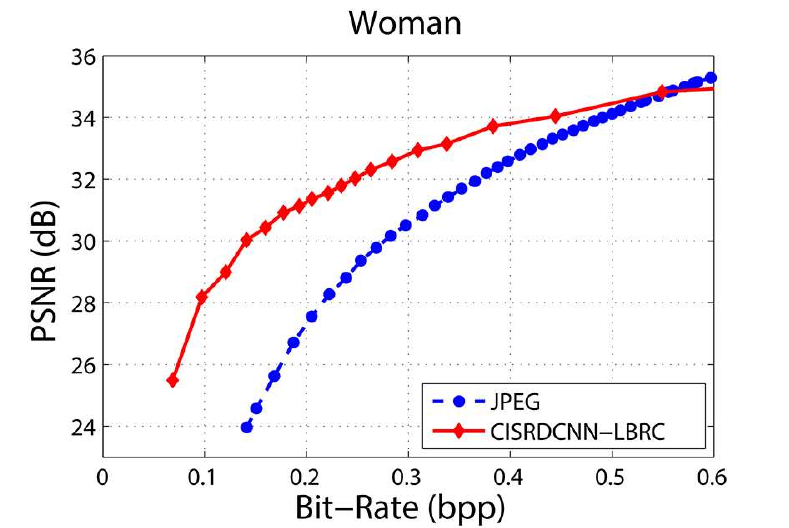}}
\subfigure[]{
\includegraphics[width=4cm]{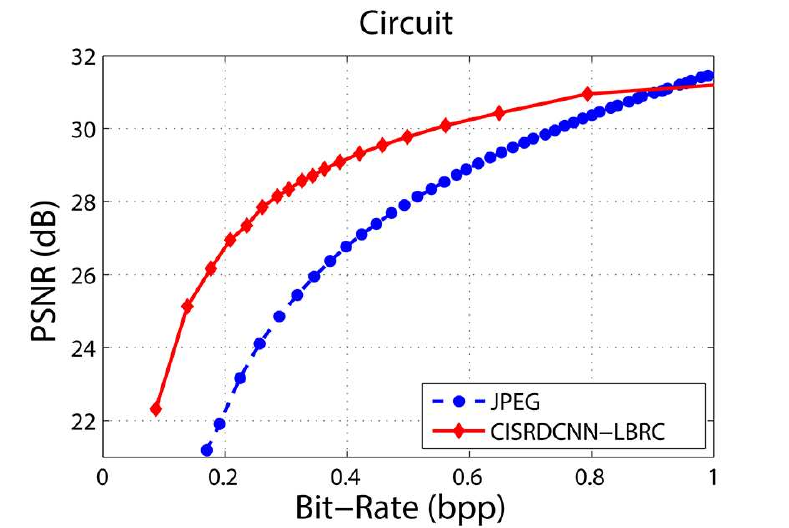}}\\
\subfigure[]{
\includegraphics[width=4cm]{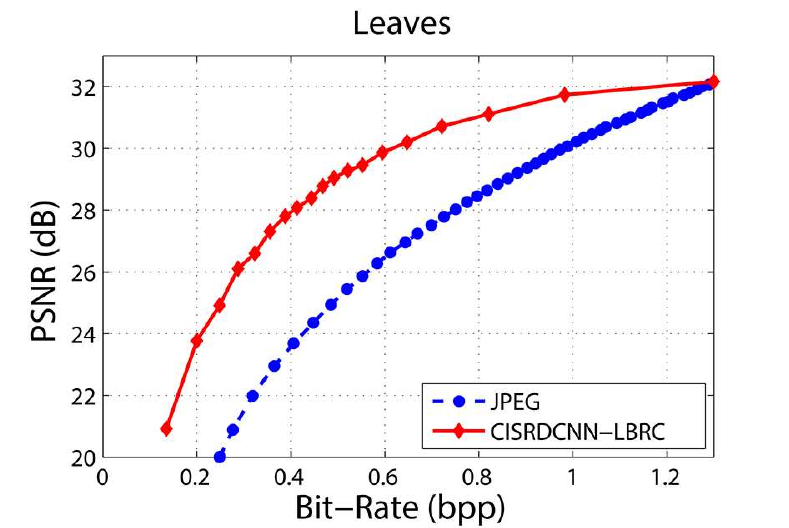}}
\subfigure[]{
\includegraphics[width=4cm]{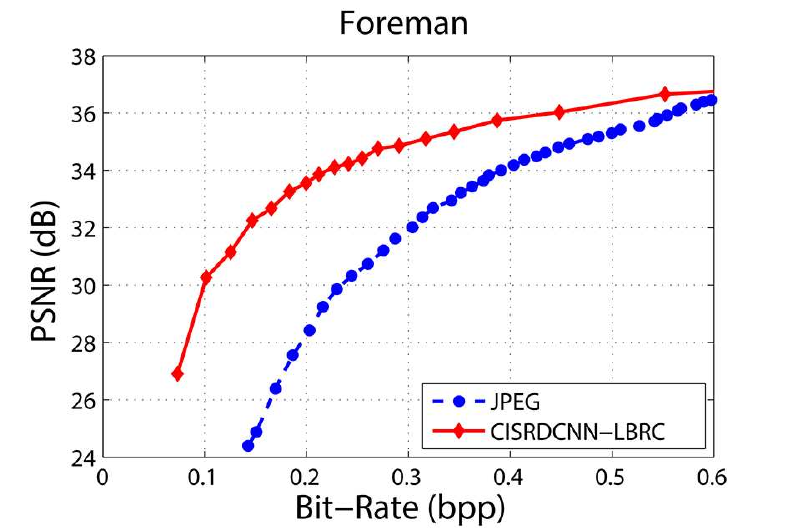}}
\subfigure[]{
\includegraphics[width=4cm]{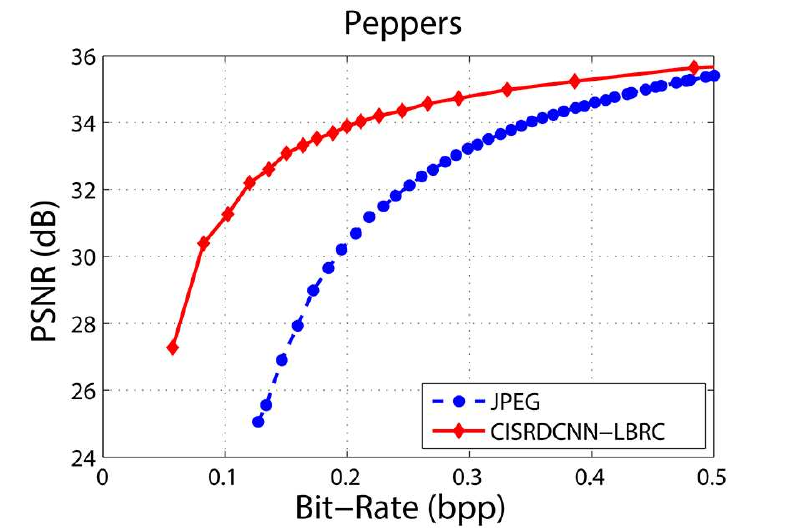}}

\caption{Rate-distortion curves of JPEG and the proposed CISRDCNN-LBRC. (a) \emph{Butterfly}. (b) \emph{Woman}. (c) \emph{Circuit}. (d) \emph{Leaves}. (e) \emph{Foreman}. (f) \emph{Peppers}.}
\label{figure.13}
\end{figure*}

\begin{figure*}[!tb]
\centering
\subfigure[]{
\includegraphics[width=2.5cm]{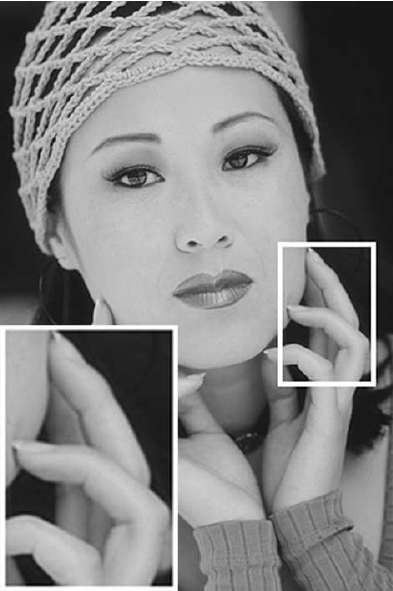}}
\subfigure[]{
\includegraphics[width=2.5cm]{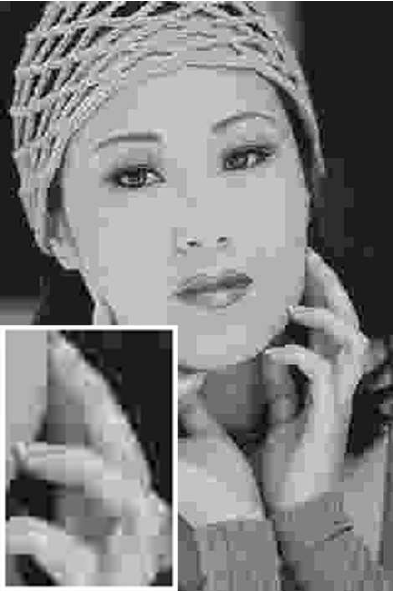}}
\subfigure[]{
\includegraphics[width=2.5cm]{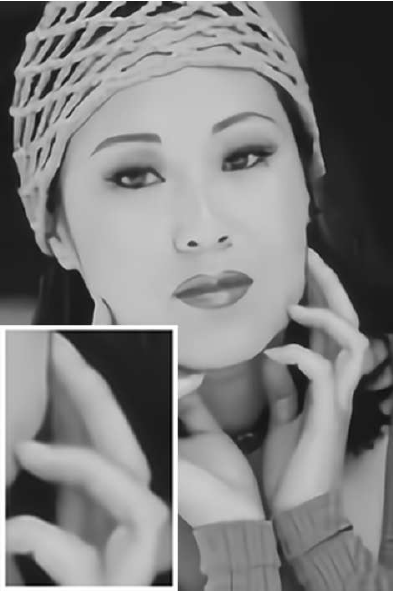}}
\subfigure[]{
\includegraphics[width=2.5cm]{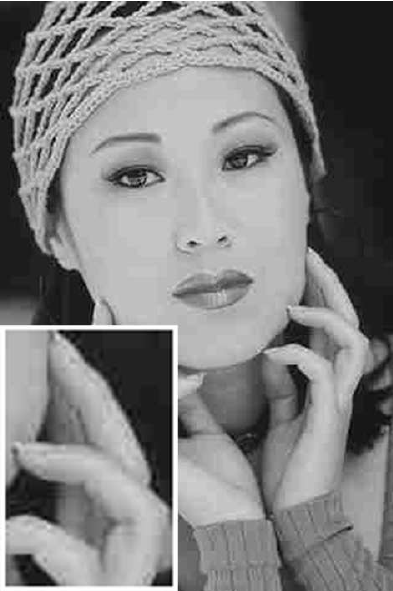}}
\subfigure[]{
\includegraphics[width=2.5cm]{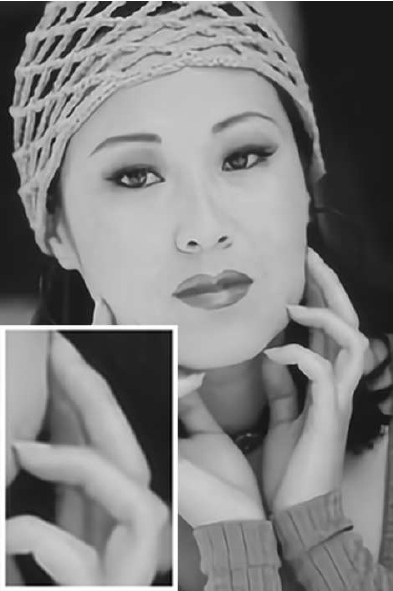}}\\

\caption{Perceptual quality comparison of JPEG and the proposed CISRDCNN-LBRC on \emph{Woman}. (a) Original image (PSNR (dB)). (b) JPEG at 0.205 bpp (27.551). (c) CISRDCNN-LBRC at 0.205 bpp (\textbf{31.359}). (d) JPEG at 0.388 bpp (32.377). (e) CISRDCNN-LBRC at 0.383 bpp (\textbf{33.710}). Please zoom in to view details and make comparisons.}
\label{figure.14}
\end{figure*}

\begin{figure*}[!tb]
\centering
\subfigure[]{
\includegraphics[width=2.5cm]{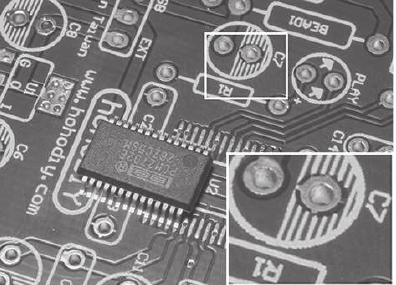}}
\subfigure[]{
\includegraphics[width=2.5cm]{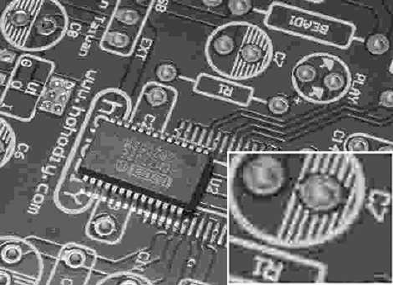}}
\subfigure[]{
\includegraphics[width=2.5cm]{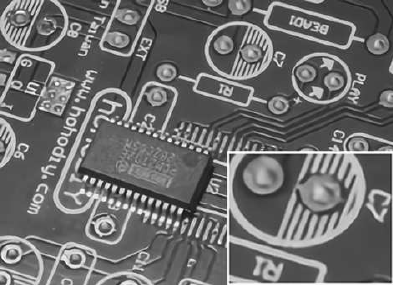}}
\subfigure[]{
\includegraphics[width=2.5cm]{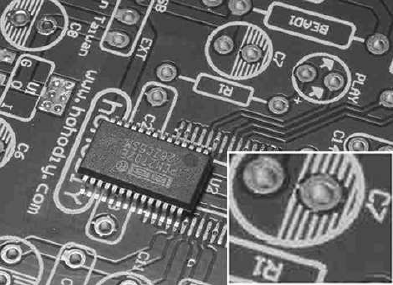}}
\subfigure[]{
\includegraphics[width=2.5cm]{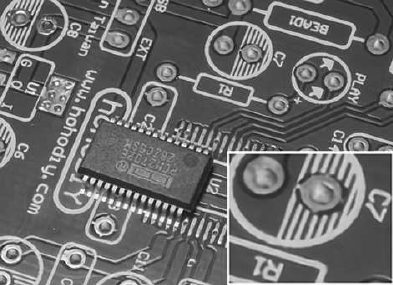}}\\

\caption{Perceptual quality comparison of JPEG and the proposed CISRDCNN-LBRC on \emph{Circuit}. (a) Original image (PSNR (dB)). (b) JPEG at 0.289 bpp (24.849). (c) CISRDCNN-LBRC at 0.286 bpp (\textbf{28.147}). (d) JPEG at 0.579 bpp (28.728). (e) CISRDCNN-LBRC at 0.561 bpp (\textbf{30.089}). Please zoom in to view details and make comparisons.}
\label{figure.15}
\end{figure*}

\begin{figure*}[!tb]
\centering
\subfigure[]{
\includegraphics[width=2.5cm]{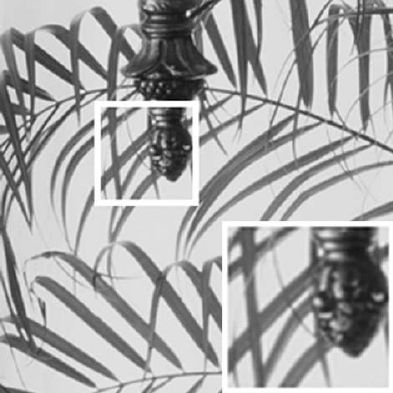}}
\subfigure[]{
\includegraphics[width=2.5cm]{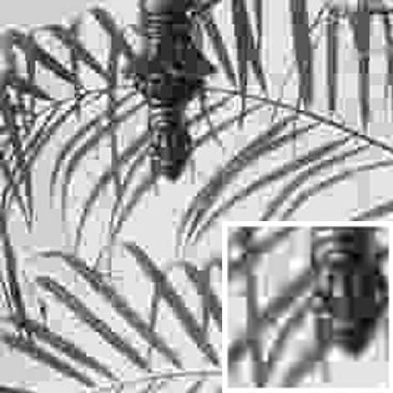}}
\subfigure[]{
\includegraphics[width=2.5cm]{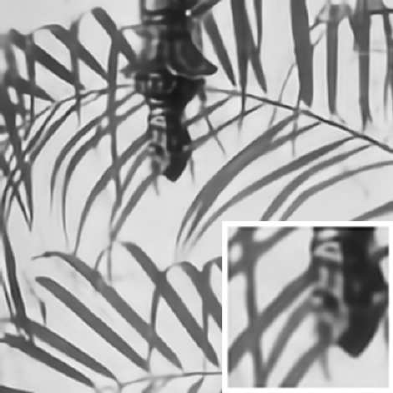}}
\subfigure[]{
\includegraphics[width=2.5cm]{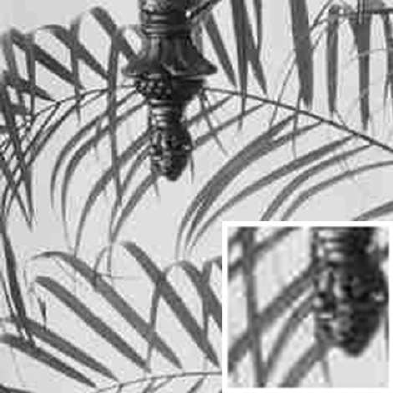}}
\subfigure[]{
\includegraphics[width=2.5cm]{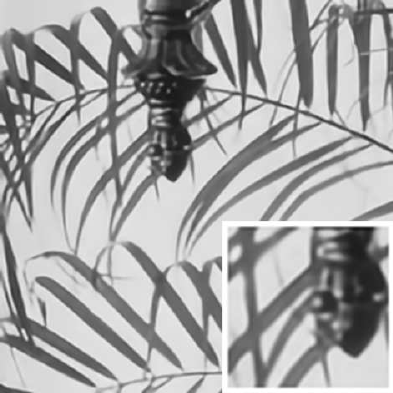}}\\

\caption{Perceptual quality comparison of JPEG and the proposed CISRDCNN-LBRC on \emph{Leaves}. (a) Original image (PSNR (dB)). (b) JPEG at 0.364 bpp (22.947). (c) CISRDCNN-LBRC at 0.355 bpp (\textbf{27.305}). (d) JPEG at 0.840 bpp (28.840). (e) CISRDCNN-LBRC at 0.821 bpp (\textbf{31.105}). Please zoom in to view details and make comparisons.}
\label{figure.16}
\end{figure*}

\begin{figure*}[!tb]
\centering
\subfigure[]{
\includegraphics[width=2.5cm]{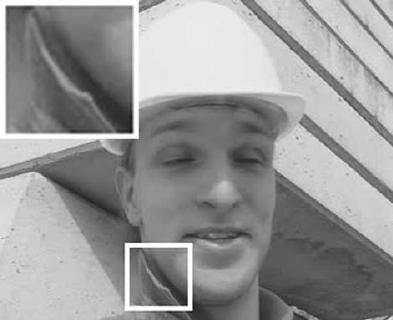}}
\subfigure[]{
\includegraphics[width=2.5cm]{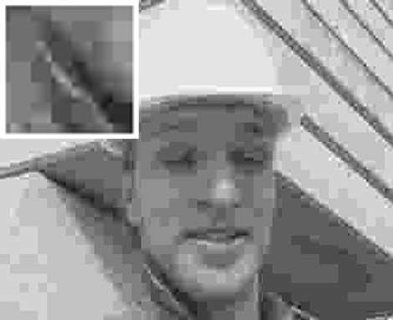}}
\subfigure[]{
\includegraphics[width=2.5cm]{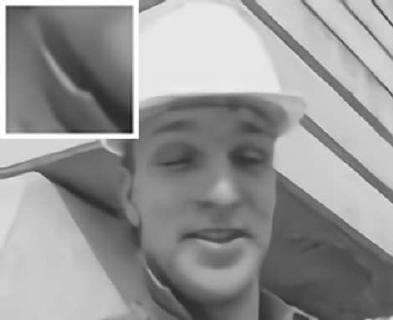}}
\subfigure[]{
\includegraphics[width=2.5cm]{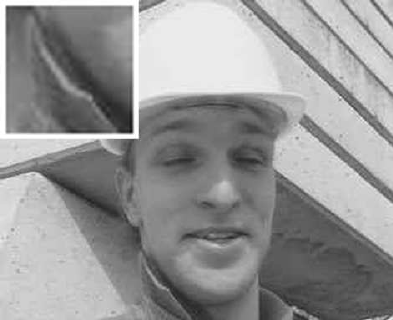}}
\subfigure[]{
\includegraphics[width=2.5cm]{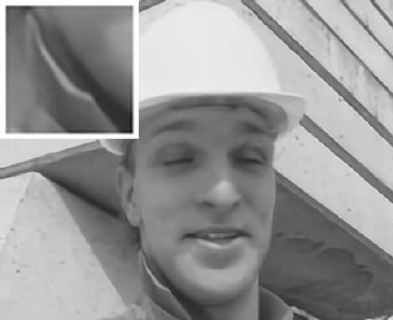}}\\

\caption{Perceptual quality comparison of JPEG and the proposed CISRDCNN-LBRC on \emph{Foreman}. (a) Original image (PSNR (dB)). (b) JPEG at 0.203 bpp (28.415). (c) CISRDCNN-LBRC at 0.200 bpp (\textbf{33.550}). (d) JPEG at 0.554 bpp (35.922). (e) CISRDCNN-LBRC at 0.552 bpp (\textbf{36.654}). Please zoom in to view details and make comparisons.}
\label{figure.17}
\end{figure*}

\subsection{Application in low bit-rate image coding}
At low bit-rates, the existing compression methods (e.g., JPEG and JPEG 2000) always produce visually unpleasant compression artifacts. In this subsection, we take the JPEG as an example to show how to use the proposed CISRDCNN to construct a low bit-rate coding framework (CISRDCNN-LBRC), thus enhancing the rate-distortion performance of JPEG. The starting point is to reduce data volume but preserve main structure of the original image via placing a downsampling operator before JPEG encoder. Correspondingly, the CISRDCNN module is placed behind the JPEG decoder to perform upsampling. As shown in Fig. \ref{figure.12}, the presented CISRDCNN-LBRC consists of four parts: downsampling, JPEG encoder, JPEG decoder, and CISRDCNN.

The test images \emph{Butterfly}, \emph{Woman}, \emph{Circuit}, \emph{Leaves}, \emph{Foreman}, and \emph{Peppers} are selected as examples to test the effectiveness of CISRDCNN-LBRC. Note that we use the luminance components of the six test images to conduct experiments in this subsection. For a fair comparison, the JPEG is used as the baseline in this experiment. The rate-distortion curves of JPEG and CISRDCNN-LBRC are presented in Figs. 13. It can be seen that the rate-distortion performance CISRDCNN-LBRC is obviously superior to the JPEG at low bit-rates. From another point of view, the CISRDCNN-LBRC can save lots of coding bits.

To compare the perceptual quality of the decoded images, we show the results of CISRDCNN-LBRC and JPEG at different bit-rates. Due to the limited space, only the results of \emph{Woman}, \emph{Circuit}, \emph{Leaves}, and \emph{Foreman} are presented in Fig. \ref{figure.14} to Fig. \ref{figure.17}. We can observe that the CISRDCNN-LBRC generates fewer artifacts and preserves main structures better. For instance, the fingers in image \emph{Woman} (Fig. \ref{figure.14}) and the collar in image \emph{Foreman} (Fig. \ref{figure.17}). Overall, the CISRDCNN-LBRC performs better than JPEG at low bit-rates in terms of both objective and subjective evaluation.

\section{Conclusion}

In this paper we propose a SR algorithm for compressed images. Unlike the existing SR methods for compressed images, we treat this task as two relevant subproblems, i.e., CAR and SR. Further, a deep network is designed to realize joint optimization of the two subproblems. We take the compression standard JPEG as an example to test the effectiveness of the proposed CISRDCNN, and experiments on both synthetic images and real low quality web images show that it produces state-of-the-art SR results. Moreover, we show the application of the proposed SR method in low bit-rate image coding, which improves the rate-distortion performance of JPEG. Intuitively, the proposed SR method and the low bit-rate coding framework can be easily extended to other image and video compression standards, e.g., JPEG 2000, H.264, and HEVC. In addition, this work provides some insights on the SR of low quality LR images (e.g., noisy and blurry), which will attract other researchers to concern this kind of problems.

However, due to the high complexity of training process and the lake of high performance computing devices, the parameters of the proposed framework are not well optimized, such as the number of layers and filters, the size of kernels, etc. In future, we will study on the settings of main parameters, which may lead to better performance and lower complexity.

\section*{Acknowledgment}
Funding: this work was supported by the National Natural Science Foundation of China [grant number 61471248]; and the National Postdoctoral Program for Innovative Talents of China [grant number BX201700163]; and the Post-Doctoral Research and Development Foundation of Sichuan University [grant number 2017SCU12003].

The authors would like to thank the authors of \cite{34,40,41,49,54,61,62,63,64} for providing their codes.

\section*{References}

\end{document}